\documentclass{article}
\usepackage{iclr2027_conference,times}
\iclrfinalcopy 

\usepackage[utf8]{inputenc} 
\usepackage[T1]{fontenc}    
\usepackage{enumitem}
\usepackage{hyperref}       
\usepackage{url}            
\usepackage{booktabs}       
\usepackage{makecell}       
\usepackage{amsfonts}       
\usepackage{nicefrac}       
\usepackage{microtype}      
\usepackage{xcolor}         
\usepackage{hyperref}
\usepackage{algorithm}  
\usepackage{algorithmic} 
\usepackage[pdftex]{graphicx}
\usepackage{epstopdf}
\usepackage{amsmath}
\usepackage{amsthm}
\usepackage{multirow}
\usepackage{tabularx}
\usepackage{caption}
\usepackage{float}
\usepackage[skins, listings, breakable]{tcolorbox}
\usepackage{listings}
\usepackage{xcolor}
\usepackage{authblk}
\usepackage{colortbl}

\usepackage[utf8]{inputenc} 
\usepackage[T1]{fontenc}    
\usepackage{hyperref}       
\usepackage{url}            
\usepackage{booktabs}       
\usepackage{amsfonts}       
\usepackage{nicefrac}       
\usepackage{microtype}      
\usepackage{xcolor}         

\usetikzlibrary{arrows.meta,positioning,calc,fit,backgrounds,shapes.geometric}

\definecolor{FigBlue}{RGB}{232,241,250}
\definecolor{FigGreen}{RGB}{232,245,236}
\definecolor{FigOrange}{RGB}{252,243,228}
\definecolor{FigRed}{RGB}{251,235,237}
\definecolor{FigGray}{RGB}{246,247,249}
\definecolor{FigPurple}{RGB}{239,235,249}

\tikzset{
figbox/.style={
  draw=black!60,
  rounded corners=3pt,
  thick,
  fill=FigGray,
  align=center,
  inner sep=5pt,
  minimum height=8mm
},
figblue/.style={figbox, fill=FigBlue},
figgreen/.style={figbox, fill=FigGreen},
figorange/.style={figbox, fill=FigOrange},
figred/.style={figbox, fill=FigRed},
figpurple/.style={figbox, fill=FigPurple},
figwhite/.style={figbox, fill=white},
figplaceholder/.style={figbox, fill=white, dashed},
figarrow/.style={-{Latex[length=2.4mm]}, line width=0.95pt, draw=black!70},
figline/.style={line width=0.9pt, draw=black!60},
treenode/.style={circle, draw=black!65, fill=white, inner sep=1.6pt},
fignote/.style={font=\scriptsize, text=black!70, align=center}
}

\usepackage{listings}
\usepackage{xcolor}
\lstdefinelanguage{json}{
    basicstyle=\normalfont\ttfamily,
    numbers=left,
    numberstyle=\scriptsize,
    stepnumber=1,
    numbersep=8pt,
    showstringspaces=false,
    breaklines=true,
    frame=lines,
    string=[s]{"}{"},
    comment=[l]{:\ "},
    morecomment=[l]{:"},
    literate=
     *{0}{{{\color{orange}0}}}{1}
      {1}{{{\color{orange}1}}}{1}
      {2}{{{\color{orange}2}}}{1}
      {3}{{{\color{orange}3}}}{1}
      {4}{{{\color{orange}4}}}{1}
      {5}{{{\color{orange}5}}}{1}
      {6}{{{\color{orange}6}}}{1}
      {7}{{{\color{orange}7}}}{1}
      {8}{{{\color{orange}8}}}{1}
      {9}{{{\color{orange}9}}}{1}
      {:}{{{\color{magenta}{:}}}}{1}
      {,}{{{\color{magenta}{,}}}}{1}
      {\{}{{{\color{blue}{\{}}}}{1}
      {\}}{{{\color{blue}{\}}}}}{1}
      {[}{{{\color{blue}{[}}}}{1}
      {]}{{{\color{blue}{]}}}}{1},
}

\PassOptionsToPackage{number,compress}{natbib}

\title{StarOR: \underline{S}ynergizing \underline{T}ree Search \underline{a}nd Test-Time \underline{R}einforcement Learning for Optimization Modeling}

\author{%
\bf
\small
Jiajun Li\textsuperscript{1},
Yu Ding\textsuperscript{1},
Shisi Guan\textsuperscript{2},
Ran Hou\textsuperscript{1},
Wanyuan Wang\textsuperscript{1,*}
\\
\small
\textsuperscript{1}School of Computer Science and Engineering, Southeast University,
\textsuperscript{2}Northwest A\&F University
\small
\textsuperscript{*}Corresponding author. E-mail: wywang@seu.edu.cn
}

\begin{document}

\maketitle
\lhead{Preprint for ICLR 2027}

\begin{abstract}
Optimization modeling is inherently hierarchical, requiring a precise sequence of symbolic commitments. Traditional learning-based automated optimization modeling methods improve modeling policies through large-scale annotated or curated training data, but are costly to adapt to new problem distributions. Meanwhile, one-shot generation remains brittle in hierarchical modeling, where early symbolic errors can propagate into invalid formulations. Test-time scaling offers a promising alternative by enabling structural exploration with additional instance-level computation; however, existing search-based methods typically rely on a fixed policy, causing repeated rollouts to inherit similar modeling biases and providing limited credit assignment for intermediate decisions. To address these limitations, we propose \textbf{StarOR}, 
a synergistic search-and-adaptation framework that couples MCTS with Test-Time Reinforcement Learning for optimization modeling. StarOR decomposes the modeling process into four stages and updates a transient LoRA adapter via GRPO at each non-terminal node. By using MCTS-generated siblings as local comparison sets, StarOR transforms search-time exploration into instance-specific policy refinement. Moreover, an unsupervised multi-faceted reward system provides fine-grained feedback for intermediate formulation decisions without ground-truth labels. Across five optimization benchmarks, StarOR attains a 65.0\% average accuracy with a 4B backbone, matching the highest average and outperforming the evaluated same-backbone test-time baselines.
\end{abstract}

\section{Introduction}
\label{intro}
Across various industries, decision-making challenges are typically articulated in natural language—ranging from complex engineering designs \citep{Belegundu_Chandrupatla_2019} to large-scale energy management \citep{krishnamurthy2018energy, singh2012overview}. While modern solvers are remarkably robust, they necessitate precise mathematical formulations to function, creating a critical translation bottleneck. This process requires an exacting sequence of logical commitments: classifying the optimization type, defining sets and parameters, and mapping constraints to executable code. Consequently, the semantic gap between a problem \emph{description} and its formal \emph{formulation} remains a primary barrier, preventing non-experts from leveraging advanced operations research tools \citep{ramamonjison2022augmenting,nl4opt2022,ahmed2024lm4opt}.

Large Language Models (LLMs) have made automated optimization modeling increasingly viable. Recent systems can already translate natural language into structured formulations or executable programs \citep{xiao2024chainofexperts, huang2024orlm, jiang2024llmopt, chen2025sirl, wu2025stepopt, wang2025ormind}. However, optimization modeling is highly sensitive to minor errors; a single silent mistake in an index or constraint direction can invalidate the entire model \citep{zhang2025sacopt, lian2026reloop}. This sensitivity stems from the task's hierarchical structure: early definitions of the problem type and sets govern all subsequent symbols, meaning errors in variables inevitably propagate to the final objective. Treating modeling as a flat, end-to-end task ignores these dependencies, making the results brittle and difficult to repair.

Current research has attempted to mitigate these issues through various methodological paradigms, yet each faces a fundamental trade-off between structural rigor and adaptive reasoning. Specifically, model-based methods using strong frontier models rely on scaling to massive parameters, which often entails prohibitive costs and privacy risks in industrial settings. In contrast, learning-based methods \citep{huang2024orlm, jiang2024llmopt, lu2025optmath, chen2025sirl} focus on refining policies offline; however, they frequently struggle with the combinatorial complexity of industrial scenarios and lack the mechanism for rapid, instance-specific iteration. Most recently, search-based methods \citep{astorga2025autoformulation, solverllm2025, liu2025optitree, wang2025bpp} enhance reliability by exploring multiple candidates at inference time. However, while search can guide traversal or candidate selection, the underlying generator remains fixed. Consequently, the process is bounded by the original proposal distribution: systematic modeling errors often recur across branches, and additional rollouts yield diminishing marginal returns—a bottleneck inherent to pure test-time scaling.

\begin{figure}[tbp]
    \centering
    \includegraphics[width=1.0\textwidth]{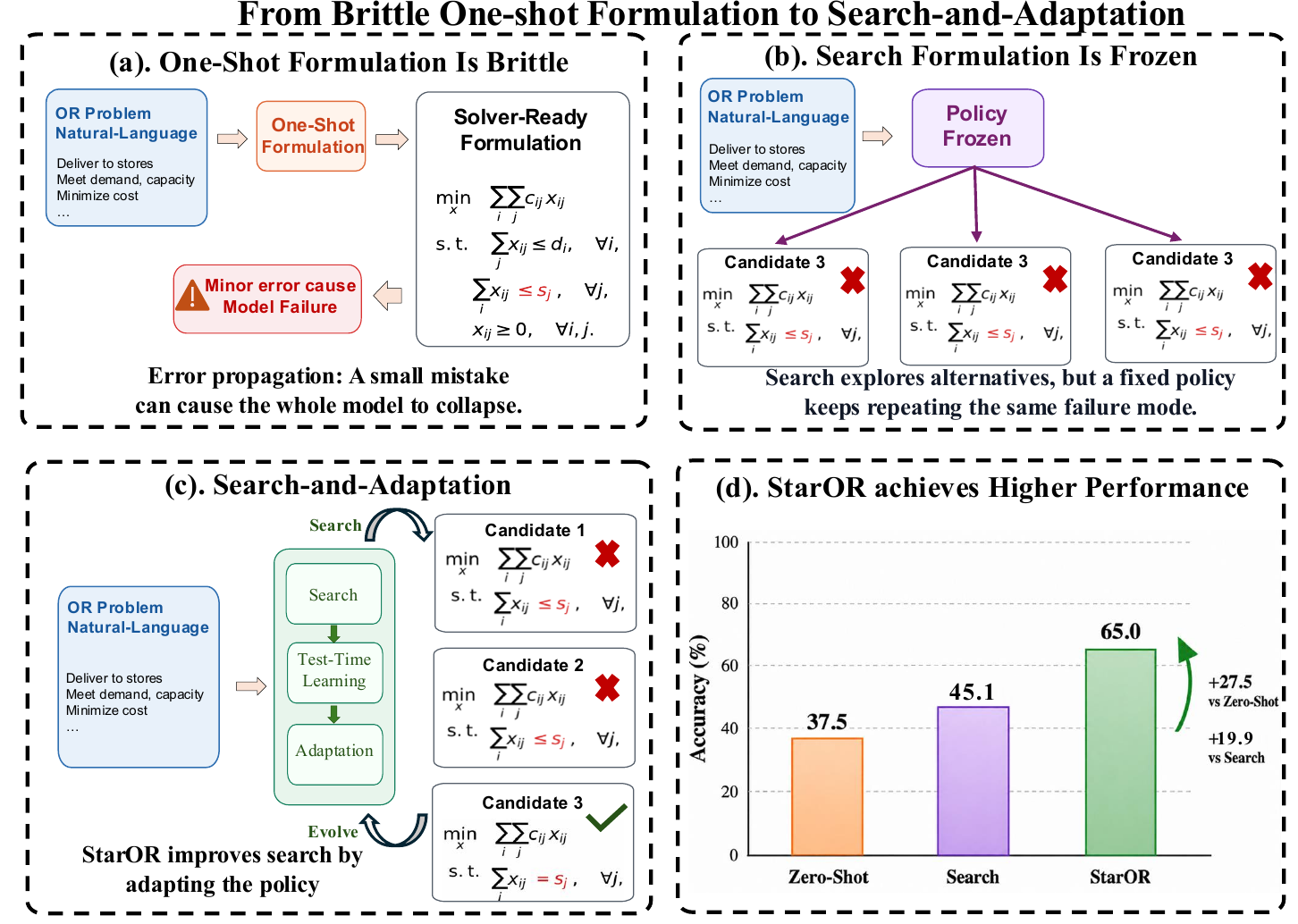}
    \caption{(a) One-shot optimization modeling is brittle because an early silent mistake can invalidate the full program. (b) Search-based methods with a fixed policy keep repeating the same failure. (c) \textsc{StarOR} searches and adapts the policy within the current instance. (d) The result empirically demonstrates the effectiveness of the search-and-adaptation paradigm.}
    \label{fig:intro}
    \vspace{-1.0em}
\end{figure}

As illustrated in Figure~\ref{fig:intro}, the limitations of existing paradigms suggest a promising direction: coupling structured exploration with instance-specific online adaptation. 
This trade-off is especially appropriate for OR modeling, where high-value industrial applications often prioritize formulation correctness over real-time generation, and additional test-time computation is acceptable when it improves reliability. 
However, realizing this coupling presents two key challenges. 
\textbf{First}, it requires a unified framework where discrete search steps serve as high-quality signals for continuous policy refinement.  
\textbf{Second}, it demands granular evaluation to address the credit assignment problem: since execution feedback is obtained only after completing partial formulations into code and ground-truth labels are unavailable at test time, the framework must infer which intermediate commitments are responsible for success or failure. 
To address these challenges, we propose \textsc{StarOR}, a framework that treats optimization modeling as a trajectory of structured commitments across four hierarchical stages. 
By anchoring a multi-faceted reward system to intermediate formulation nodes, \textsc{StarOR} enables each search step to both explore the formulation space and provide fine-grained feedback for adapting the policy to the current instance. 
This transforms optimization modeling from a static generation into a dynamic evolution. Our main contributions are: \textbf{(1) Search and adaptation:} We introduce \textsc{StarOR}, the first architecture to integrate hierarchical MCTS with test-time training for OR modeling, enabling online policy evolution during the optimization modeling process; \textbf{(2) OR-specific reward:} we propose an unsupervised multi-reward system to evaluate formulation quality without ground-truth labels; and \textbf{(3) Performance:} with a 4B backbone, \textsc{StarOR} attains a $65.0\%$ average accuracy across five benchmarks, matching the highest average and exceeding the evaluated same-backbone test-time baselines.

\section{Related Work}
\label{related}

\paragraph{Learning-based Methods for Optimization Modeling.}
LLM-based optimization modeling has progressed from benchmark construction \citep{ramamonjison2022augmenting,nl4opt2022,ahmed2024lm4opt} to systems trained with synthetic formulations and solver feedback. ORLM and OptMATH build large-scale synthesis pipelines \citep{huang2024orlm,lu2025optmath}, LLMOPT studies standardized formulation representations \citep{jiang2024llmopt}, and SIRL uses solver-informed rewards for executable modeling \citep{chen2025sirl}. Recent systems such as Step-Opt-Instruct and ORMind further emphasize structured validation and iterative reasoning \citep{wu2025stepopt,wang2025ormind}. These methods improve the base modeling prior, but they remain largely offline: adapting to new industrial constraints typically requires additional data, retraining, or stronger proprietary models.

\paragraph{Search-based Methods for Optimization Modeling.}
Search-based inference addresses the brittleness of one-shot generation by exploring multiple formulation trajectories. Inspired by Tree of Thoughts \citep{yao2023tree}, OR-specific methods use structured search to improve candidate selection: AutoFormulation combines MCTS with symbolic pruning \citep{astorga2025autoformulation}, SolverLLM uses feedback-guided dynamic expansion \citep{solverllm2025}, and OptiTree and related methods decompose modeling through hierarchical search \citep{liu2025optitree,wang2025bpp}. However, these approaches keep the policy fixed. Search can filter or rerank candidates, but it does not internalize the feedback encountered during exploration, so modeling biases may recur across branches.

\paragraph{Test-time Adaptation and Reinforcement Learning.}
Test-time training and reinforcement learning update a model or lightweight adapter during inference, allowing feedback from the current instance to shape later predictions \citep{akyurek2025surprising,zuo2025ttrl,hu2025tlm}. Related work strengthens this idea with tool verification, transient policy adaptation, and online discovery \citep{liao2026tool,jiao2026policy,wang2025thetaevolve,yuksekgonul2026learning}. In operations research, OR-R1 applies group relative policy optimization to improve modeling outputs at test time \citep{ding2025orr1}. \textsc{StarOR} follows this direction but changes the granularity of adaptation: instead of learning only from complete outputs, it attaches executable, semantic, and structural rewards to intermediate MCTS nodes, enabling instance-specific policy refinement during formulation search.

\section{Method}
\label{method}

\subsection{Problem Formulation}

Given a natural language optimization problem $x$, our objective is to generate executable, faithful, and mathematically sound Python code $c$. Following \citep{jiang2024llmopt, solverllm2025}, we decompose the modeling process into six elements—\emph{problem type, sets, parameters, variables, objective,} and \emph{constraints}—structured as a four-stage trajectory $\tau = (z^{(1)}, z^{(2)}, z^{(3)}, z^{(4)})$. These stages represent: (1) problem type and sets, (2) parameters and variables, (3) objective and constraints, and (4) the final executable code. Any prefix $\tau_{\le s}$ ($s < 4$) denotes a \emph{partial formulation}.

We formalize this sequence as a Markov Decision Process (MDP) defined by $(\mathcal{S}, \mathcal{A}, \mathcal{P}, \mathcal{R})$:
\begin{itemize}
    \item \textbf{States $\mathcal{S}$}: A state $\sigma_t = (x, \tau_{\le t})$ pairs the problem description $x$ with the partial formulation $\tau_{\le t}$ generated so far.
    \item \textbf{Actions $\mathcal{A}$}: An action $a_t$ corresponds to generating the next-stage commitment $z^{(t+1)}$.
    \item \textbf{Transitions $\mathcal{P}$}: The transition is deterministic, defined by $\sigma_{t+1} = (x, \tau_{\le t+1})$, where $\tau_{\le t+1}$ is the result of appending $a_t$ to $\tau_{\le t}$.
    \item \textbf{Rewards $\mathcal{R}$}: The task-level reward is defined on completed code at $\sigma_4$. StarOR estimates a node-level reward by completing each partial trajectory and evaluating it (see Section~\ref{reward}).
\end{itemize}

\begin{figure}[tbp]
    \centering
    \includegraphics[width=0.97\textwidth]{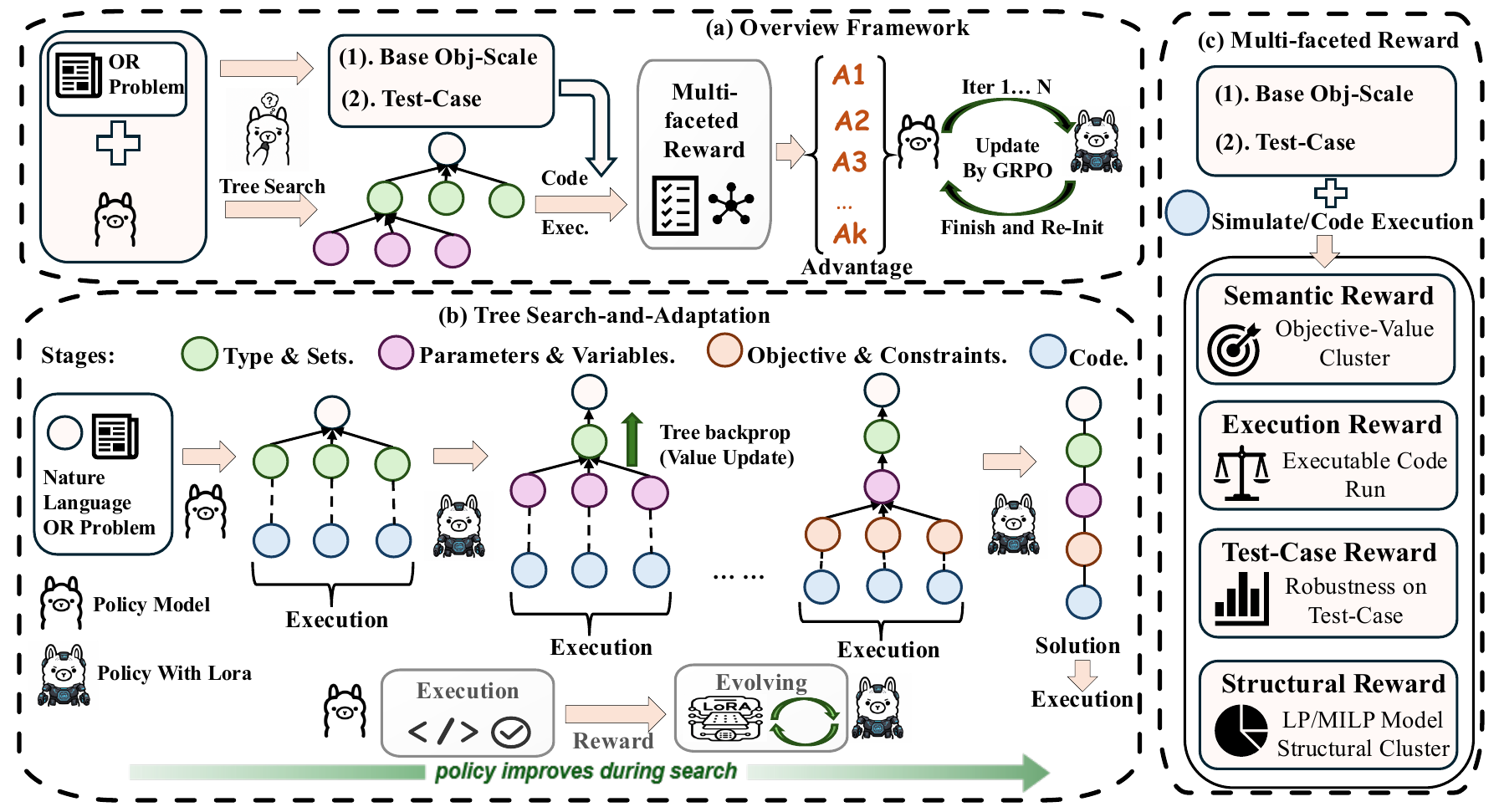}
    \caption{Overview of the StarOR framework.(a) Overall framework: StarOR forms a closed loop between tree search, code execution, multi-faceted reward evaluation, and GRPO-based transient LoRA adaptation. (b) search-and-adaptation: StarOR searches over four formulation stages and uses execution-grounded feedback to adapt the local policy during exploration. (c) Multi-faceted reward: An unsupervised reward provides node-level feedback for both tree search and policy evolution.}
    \label{fig:framework}
    \vspace{-1.0em}
\end{figure}

\subsection{StarOR: Online Optimization and Tree Search for Modeling}
\label{sec:StarOR-overview}

We propose \textit{StarOR}, a framework that transforms optimization modeling into a synergized process of Monte Carlo Tree Search (MCTS) and online policy optimization. Unlike conventional static generation, StarOR treats each modeling stage as a localized task of search-and-adaptation. Figure~\ref{fig:framework} shows the overview of the StarOR's pipeline for Optimization Modeling.

At each iteration $t$, MCTS identifies a partial formulation $\tau_{\le s-1}$ and addresses a localized sub-problem: determining the optimal next-stage component $z^{(s)}$ conditioned on the task $x$ and preceding modeling decisions. Formally, for stage $s$, we sample a candidate set:
\begin{equation}
z_i^{(s)} \sim \pi_{\phi+\Delta\phi}\!\left(\cdot \mid x, \tau_{\le s-1}, s\right), \qquad i=1,\dots,K,
\end{equation}
where $\phi$ denotes the base policy and $\Delta\phi$ a transient LoRA adapter used only for this instance.
This integration fosters a self-evolving search process: MCTS facilitates structured exploration of modeling hypotheses, while online optimization distills execution feedback to refine the transient adapter $\Delta\phi$. In this dual-purpose cycle, the sampled candidates simultaneously expand the search tree and serve as the training batch for test-time adaptation. Consequently, the framework identifies critical decision points and immediately sharpens the local policy for the specific modeling instance. The remainder of this section is organized as follows: Section~\ref{sec:policy-evolution} describes our node-level online policy evolution, while Section~\ref{sec:tree-search} provides a detailed exposition of the MCTS framework.

\subsubsection{Policy Evolution via Group Relative Policy Optimization (GRPO)}
\label{sec:policy-evolution}

\paragraph{Group Advantage Estimation via GRPO.} 
At each MCTS iteration $t$ corresponding to stage $s$, we sample a sibling group of $K$ candidate continuations 
$\{z_i^{(s)}\}_{i=1}^K$ conditioned on the same prefix $\tau_{\le s-1}$. 
Following Group Relative Policy Optimization (GRPO), these siblings serve as a local comparison group, allowing StarOR to estimate candidate quality by relative reward rather than an additional value function \citep{shao2024deepseekmath}:
\begin{equation}
A_i = \frac{R_i - \operatorname{mean}_{j=1}^{K} R_j}
{\operatorname{std}_{j=1}^{K} R_j + \epsilon},
\end{equation}
where $\epsilon$ is a small numerical constant. GRPO updates pre-code reasoning and current-stage formulation using rewards from executable completions (Appendix~\ref{app:details}).

\paragraph{Multi-faceted Reward.} \label{reward}
To stabilize the search process and prevent extreme outliers from distorting the learning signal, we first use the base policy to establish a pre-generation prior before search-and-adaptation. 
Conditioned on the problem context, the base policy performs problem-level reasoning to construct synthetic perturbation cases and estimate a plausible objective range $[L_x, U_x]$ for instance $x$. 
This prior is used only as a soft grounding signal: it calibrates the objective-range penalty in $r_{\mathrm{sem}}$ and provides expected objective ranges for $r_{\mathrm{test}}$, rather than serving as a hard correctness oracle. 
For candidate $i$, the total reward $R_i = \sum_j w_j r_{j,i}$ is computed from the following components:

\noindent \textbf{Semantic Reward ($r_{sem,i}$)}:
This measures semantic consensus via the objective-value cluster size $|C_{sem,i}|$. Let $K_{\mathrm{valid}}$ denote the number of candidates in the sibling rollout group that are executable and return a finite objective value. We cluster these valid objective values under a numerical tolerance, and $C_{sem,i}$ denotes the cluster containing candidate $i$. To penalize numerically implausible results while maintaining robustness against potential inaccuracies in the initial estimation, we apply a soft-penalty factor $\lambda \in (0,1)$ for objectives falling outside the predicted range:
\begin{equation} \label{reward_sem}
    r_{sem,i} = \frac{|C_{sem,i}|}{K_{\mathrm{valid}}} \cdot 
    \begin{cases} 
    1 & \text{if } O_i \in [L_x, U_x], \\
    \lambda & \text{otherwise.}
    \end{cases}
\end{equation}
For candidates that fail to execute or do not produce a finite objective value, we set
$r_{sem, i}=0$. This $\lambda$-scaling ensures that otherwise valid formulations are not
entirely dismissed due to conservative estimates from the pre-generation phase.

\noindent \textbf{Structural Reward ($r_{struct,i}$)}: This evaluates structural consistency via a signature vector $\mathbf{v}_i = [n_{\mathrm{bin}}, n_{\mathrm{int}}, n_{\mathrm{cont}}, n_{\mathrm{con}}, \mathrm{sense}]$, representing the counts of binary, integer, and continuous variables, the number of constraints, and the optimization sense. Let $|C(i,d)|$ denote the number of executable candidates sharing candidate $i$'s value on dimension $d$. The reward is:
\begin{equation} \label{reward_struct}
    r_{struct,i} = \frac{1}{5} \sum_{d \in \mathbf{v}_i}
    \sqrt{\frac{|C(i,d)|}{K^*_{\mathrm{valid}}}}.
\end{equation}
Here $K^*_{\mathrm{valid}}$ is the number of executable candidates.

\noindent \textbf{Execution Reward ($r_{exec,i}$)}: $\in \{0, 1\}$, indicating code executability.

\noindent \textbf{Test-case Reward ($r_{test,i}$)}: This provides grounding through $N_t$ test cases. Each rollout is assigned a score $S_{i,j} \in \{0, \lambda, 1\}$ based on whether its output aligns with the expected range for test case $j$:
\begin{equation} \label{reward_test}
    r_{test,i} = \frac{1}{N_t} \sum_{j=1}^{N_t} S_{i,j}.
\end{equation}

To optimize the search-and-adapt process, \textbf{Dynamic Reward Shaping} adaptively tunes the weights $w_j$ as MCTS iterations progress. It transitions the learning signal from \emph{feasibility-oriented} metrics ($r_{exec}, r_{test}$) to \emph{consensus-oriented} metrics ($r_{sem}, r_{struct}$), thereby balancing broad initial exploration with eventual convergence toward a precise and consistent mathematical architecture. A detailed description for reward is provided in the Appendix~\ref{app:reward}.

\paragraph{Sample-Specific Policy Evolution.} 
This reward-driven feedback is internalized through a transient LoRA adapter $\Delta\phi$, which is initialized to zero and reset for every new problem instance to prevent cross-task interference. This mechanism facilitates a policy evolution where search and learning mutually reinforce one another: MCTS generates the local contrastive hypotheses necessary for meaningful advantage estimation, while GRPO distills the execution feedback to sharpen the policy's reasoning for the specific instance. This coupling overcomes the limitations of static exploration; while MCTS identifies alternative modeling paths, the online optimization ensures that the policy progressively specializes in the unique constraints and nuances of the current OR problem, refining the model from high-level set definitions down to terminal code implementation.

\subsubsection{Monte Carlo Tree Search for Optimization Modeling}
\label{sec:tree-search}

We formulate the modeling process as a structured decision-making task, navigating a four-stage search tree via Monte Carlo Tree Search (MCTS). This framework systematically explores the space of LLM-proposed formulations through four refined phases: \texttt{Problem Type and Sets}, \texttt{Parameters and Variables}, \texttt{Objective and Constraints}, and \texttt{Code}.

\paragraph{Selection.}
The search traverses the tree from the root by selecting child nodes that maximize a PUCT-style objective, balancing exploitation and exploration:
\begin{equation}
\mathrm{Score}(n, a) = Q(n, a) + c_{\mathrm{puct}} \cdot P(n, a) \frac{\sqrt{\sum_{a'} N(n, a')}}{1 + N(n, a)},
\end{equation}
where $n$ denotes the current node and the prior probability $P(n, a)$ is derived from the average log-likelihood of the sequence $Y_a$ corresponding to action $a$:
\begin{equation}
P(n, a) = \frac{\exp(s_a / \eta)}{\sum_{a' \in \mathcal{A}(n)} \exp(s_{a'} / \eta)}, \quad 
s_a = \frac{1}{|Y_a|} \sum_{u=1}^{|Y_a|} \log \pi_{\theta}(y_u \mid x, \tau_{\leq s-1}, y_{<u}).
\end{equation}
Here, $\eta$ is the prior softmax temperature, and $\pi_{\theta}$ (where $\theta = \phi + \Delta\phi$) represents the transient policy. This mechanism ensures that the selection is guided by both historical rewards $Q$ and the model's evolving, instance-specific beliefs.

\paragraph{Expansion and Simulation.}
Upon reaching an expandable leaf node at stage $s$, StarOR expands the tree by sampling $K$ candidate commitments $\{z_i^{(s)}\}_{i=1}^K$ from the current transient policy $\pi_{\phi+\Delta\phi_t}$. Each candidate is appended to the existing partial formulation and temporarily completed into executable code for evaluation (In a single generation). After execution, sibling candidates that yield near-identical objective values (tolerance $\epsilon_c=0.01\%$) are aggregated into a \textit{Group Cluster}, representing behaviorally similar modeling paths under the current instance. This clustering enables the search to share evidence across semantically equivalent branches and reduce redundant exploration.

\paragraph{Group Backpropagation.}
To leverage collective evidence during tree search, we propagate rewards at both individual and group granularities. 
Each child node $i$ maintains its own reward $R_i$, while the selected ancestral path is updated using the group mean 
$\bar{R} = \frac{1}{K}\sum_{i=1}^{K} R_i$. 
Beyond standard backpropagation, we introduce \textit{Group Backpropagation} to share evidence across behaviorally similar nodes. 
The intuition is that nodes within the same Group Cluster often correspond to formulations with similar objective behavior and hence similar values; repeatedly revisiting them therefore leads to redundant exploration. 
For any sibling node $m$ residing in the same Group Cluster as a trajectory node, its value is softly updated as:
\begin{equation}
Q(m) \leftarrow \frac{N(m) Q(m) + \rho^{\ell} \bar{R}}{N(m) + \rho^{\ell}},
\end{equation}
where $N(m)$ is incremented by $\rho^{\ell}$, $\ell$ denotes the structural tree distance between node $m$ and the trajectory node, and $\rho \in [0,1]$ is a decay factor. 
This shared update allows PUCT to treat same-cluster nodes as partially explored, reducing repeated visits to semantically equivalent branches.

\paragraph{Early Stopping and Termination.}
The search terminates upon reaching the \texttt{code} stage. To enhance symbolic coverage, StarOR applies a one-shot suppression factor $\gamma_{\mathrm{sup}}=0.5$ to the PUCT scores of the current trajectory and its associated objective-consensus cluster. This temporarily diverts the search from the current basin, forcing the exploration of alternative branches before final acceptance. Early stopping is triggered if three consecutive steps across distinct stages converge to the same objective consensus. Upon termination, the optimal candidate is selected from the \texttt{code} nodes; if the budget is exhausted, a global consensus voting mechanism identifies the most robust formulation. Any residual errors trigger a repair loop guided by execution traces. This synergy allows MCTS to navigate the formulation space while the repair phase finalizes executable reliability.

\section{Experiments}
\label{experiments}

\definecolor{ExpHeader}{RGB}{242,245,249}
\definecolor{ExpBand}{RGB}{246,248,251}
\definecolor{ExpAccent}{RGB}{231,244,236}
\newcommand{\resultpending}{\textcolor{black!55}{\textit{TBD}}}
\newcommand{\resultna}{\textcolor{black!45}{--}}
\newcommand{\benchcell}[2]{\makecell[c]{\scriptsize\textbf{#1}\\[-1pt]\scriptsize\textbf{#2}}}
\newcommand{\benchsingle}[1]{\makecell[c]{\scriptsize\textbf{#1}}}
\newcommand{\subvariant}[1]{\hspace{0.35em}\textit{#1}}

To evaluate the effectiveness of StarOR, we conduct comprehensive experiments across five
benchmark datasets, comparing our approach with both prompt-based and learning-based baselines.
Our study is designed to answer the following key research questions:

\begin{enumerate}[label=\textbf{(RQ\arabic*)}, leftmargin=3.5em, labelsep=0.5em, itemsep=0.5ex]
    \item \textbf{Effectiveness of StarOR:} How does our proposed StarOR, as a test-time scalable framework, compare with  Strong large models (e.g., GPT-4, DeepSeek-R1), specialized learning-based model and some test-time reasoning method across diverse optimization benchmarks?
    
    \item \textbf{Synergy of Search and Test-Time RL:} What is the specific contribution of Test-Time Adaptation to the overall performance? How does the framework perform when utilizing only MCTS without the iterative GRPO-based policy update?
    
    \item \textbf{Impact of Reward Mechanism:} How does our specialized reward design influence the search quality and decision-making during inference? Can it significantly outperform simpler baselines such as majority voting?
    
    \item \textbf{Test-Time Scalability:} How does the performance of StarOR scale with respect to the expansion group size $K$ and the overall computational budget?
\end{enumerate}

\subsection{Experiment Setup}
\paragraph{Benchmarks.}
We evaluate on five cleaned benchmarks totaling 1356 instances following \cite{chen2025sirl}. Table~\ref{tab:benchmarks} summarizes the main benchmarks.

\begin{table}[H]
\centering
\small
\setlength{\tabcolsep}{5pt}
\renewcommand{\arraystretch}{1.12}
\vspace{-0.5em}
\caption{Main benchmark suite. \#Inst. is the number of instances in our evaluation files. The benchmarks are roughly ordered by increasing modeling difficulty from top to bottom.}
\label{tab:benchmarks}
\begin{tabularx}{\columnwidth}{>{\bfseries}l>{\centering\arraybackslash}p{0.14\columnwidth}X}
\toprule
\rowcolor{ExpHeader}
Dataset & \#Inst. & Focus \\
\midrule
NL4OPT (\cite{nl4opt2022})  & 245 & Natural-language OR modeling \\
MAMO-Easy (\cite{huang2024mamo}) & 642 & Easy LP/MILP modeling \\
MAMO-Complex (\cite{huang2024mamo})& 203 & Complex LP/MILP modeling \\
IndustryOR (\cite{huang2024orlm})& 100 & Industrial OR problems \\
OptMATH (\cite{lu2025optmath})& 166 & Optimization modeling benchmark \\
\midrule
\rowcolor{ExpBand}
Total & 1356 &  Varying-difficulty OR modeling tasks \\
\bottomrule
\vspace{-0.5em}
\end{tabularx}
\end{table}

\paragraph{Baselines.} To benchmark our method, we compare against three categories of baseline approaches (see Appendix~\ref{app:baselines} for detailed descriptions):

\begin{enumerate}[label=\arabic*., leftmargin=1.5em, labelsep=0.5em, nosep, topsep=2pt]
    \item \textbf{Model-based Methods.} Approaches that utilize state-of-the-art frontier models, such as GPT-4 \citep{achiam2023gpt4}, DeepSeek-V3.1 \citep{deepseekai2024deepseekv3}, and Large Reasoning Model DeepSeek-R1 \citep{guo2025deepseekr1}, OpenAI-o3 \citep{openai2025o3}, integrated with the Gurobi solver \citep{gurobi} for optimization tasks.
    
    \item \textbf{Learning-based Methods.} Methods that enhance the backbone model through offline training or synthetic supervision prior to inference. This category includes ORLM \citep{huang2024orlm}, LLMOPT \citep{jiang2024llmopt}, OptMATH \citep{lu2025optmath}, and SIRL \citep{chen2025sirl}.
    
    \item \textbf{Test-time Methods.} Strategies that operate during inference to refine solution quality using the same backbone when applicable. We compare against zero-shot decoding, best-of-$N$ \citep{kang2025scalable}, Reflexion \citep{shinn2023reflexion}, AutoFormulator (MCTS-Style method) \citep{astorga2025autoformulation}, OptiTree \citep{liu2025optitree}, and OR-R1 \citep{ding2025orr1}.
\end{enumerate}

\paragraph{Implementation Details and Metrics.} To ensure a fair comparison, we standardize on Qwen3-4B-Instruct-2507 as the backbone for test-time methods, except OR-R1 (Qwen3-8B). Detailed hyperparameters are provided in Appendix \ref{app:hyper}. We adopt \textit{Solution Accuracy} (\textbf{Acc.}) as our primary metric, using a numerical tolerance of $10^{-6}$ for verification. Moreover, we use expansion group size $K=8$ and max iterations $T =10$ to conduct experiment.

\subsection{Result Analysis}
\paragraph{Main results (RQ1).}
As shown in Table~\ref{tab:main-results}, \textsc{StarOR}-4B attains a 65.0\% average across five benchmarks, matching DeepSeek-V3.1 and exceeding DeepSeek-R1 and GPT-4 in macro average. It improves on SIRL-7B's 60.6\% average by 4.4 percentage points and outperforms the evaluated same-backbone test-time baselines across all five benchmarks. On \textit{IndustryOR}, \textsc{StarOR} reaches the highest accuracy of 51.0\%. These results show that instance-level search and adaptation can deliver strong optimization-modeling accuracy with a 4B backbone.

\begin{table}[H]
\centering
\footnotesize 
\setlength{\tabcolsep}{2.8pt}
\renewcommand{\arraystretch}{1.18}
\vspace{-0.5em}
\caption{Main performance comparison (Acc.\ \%) on five benchmarks. Macro Avg. is the unweighted mean of the five benchmark accuracies. Results marked with $*$ are reported in \cite{chen2025sirl}, except for OR-R1, whose results are taken from its original paper. $-$ denotes missing results.}
\label{tab:main-results}
\begin{tabularx}{\columnwidth}{>{\raggedright\arraybackslash}p{3.35cm}@{\hspace{3pt}}*{5}{>{\centering\arraybackslash}X}@{\hspace{2pt}\color{black!18}\vrule width 0.5pt\hspace{4pt}}>{\columncolor{ExpHeader!70}\centering\arraybackslash}X}
\toprule
\rowcolor{ExpHeader}
\makecell[l]{\rule{0pt}{2.2ex}\textbf{Method}} & \benchsingle{NL4OPT} & \benchcell{MAMO}{Easy} & \benchcell{MAMO}{Complex} & \benchcell{Industry}{OR} & \benchsingle{OptMATH} & \benchcell{Macro}{Avg.} \\
\midrule
\rowcolor{ExpBand}
\multicolumn{7}{l}{\textbf{Model-based Methods}} \\
GPT-4 & 89.0$^*$ & 87.3$^*$ & 49.3$^*$ & 33.0$^*$ & 16.6$^*$ & 55.0$^*$ \\
DeepSeek-V3.1 & 84.8$^*$ & 88.9$^*$ & 63.5$^*$ & 44.0$^*$ & 43.9$^*$ & 65.0$^*$ \\
DeepSeek-R1 & 82.4$^*$ & 87.2$^*$ & 67.9$^*$ & 45.0$^*$ & 40.4$^*$ & 64.6$^*$ \\
OpenAI-o3 & 69.4$^*$ & 77.1$^*$ & 51.2$^*$ & 44.0$^*$ & 44.0$^*$ & 57.1$^*$ \\
\addlinespace[2pt]
\rowcolor{ExpBand}
\multicolumn{7}{l}{\textbf{Learning-based Methods}} \\
ORLM-LLaMA3-8B & 85.7$^*$ & 82.3$^*$ & 37.4$^*$ & 24.0$^*$ & 2.6$^*$ & 46.4$^*$ \\
LLMOPT-Qwen2.5-14B & 80.3$^*$ & 89.5$^*$ & 44.1$^*$ & 29.0$^*$ & 12.5$^*$ & 51.1$^*$ \\
OptMATH-Qwen2.5-7B & 94.7$^*$ & 86.5$^*$ & 51.2$^*$ & 20.0$^*$ & 24.4$^*$ & 55.4$^*$ \\
SIRL-Qwen2.5-7B & 96.3$^*$ & 91.7$^*$ & 51.7$^*$ & 33.0$^*$ & 30.5$^*$ & 60.6$^*$ \\
\addlinespace[2pt]
\rowcolor{ExpBand}
\multicolumn{1}{l}{\textbf{Test-Time Methods}} & \multicolumn{6}{c}{\footnotesize\textit{Base model: Qwen3-4B-Instruct-2507}} \\
Zero-shot & 55.9 & 81.9 & 11.8 & 27.0 & 10.8 & 37.5 \\
Best-of-$N$($N=16$) & 70.6 & 88.9 & 22.2 & 39.0 & 21.7 & 48.5 \\
Reflexion & 58.0 & 82.7 & 16.3 & 28.0 & 11.4 & 39.3 \\
AutoFormulator & 75.5 & 85.7 & 23.2 & 30.0 & 11.4 & 45.2 \\
OptiTree & 86.5 & 90.0 & 32.0 & 33.0 & 12.7 & 50.8 \\
OR-R1-Qwen3-8B & 88.3$^*$ & 86.1$^*$ & 49.9$^*$ & 35.3$^*$ & -- & -- \\
\rowcolor{ExpAccent}
\textbf{StarOR (Ours)} & \normalsize\textbf{92.7} & \normalsize\textbf{94.4} & \normalsize\textbf{54.2} & \normalsize\textbf{51.0} & \normalsize\textbf{32.5} & \normalsize\textbf{65.0} \\
\bottomrule
\end{tabularx}
\end{table}

The OptMATH Best-of-$N$ ($N=16$) entry in this table is a single-run result; Table~\ref{tab:scaling} reports the three-seed mean, with details in Table~\ref{tab:bon-multiseed}.

\paragraph{Effect of LoRA Adaptation on Search (RQ2).}
Table~\ref{tab:rq2-search-vs-ttrl} compares search-only exploration with the full StarOR framework to assess the effect of LoRA adaptation. While search alone already improves over the zero-shot baseline, online LoRA adaptation brings consistent further gains across benchmarks. This shows that StarOR can internalize search-time feedback through policy updates, moving beyond pure exploration toward more robust reasoning under complex constraints.

\begin{table}[H]
\centering
\footnotesize
\setlength{\tabcolsep}{2.8pt}
\renewcommand{\arraystretch}{1.1}
\vspace{-0.5em}
\caption{(RQ2) Contribution of online policy optimization beyond pure search.}
\label{tab:rq2-search-vs-ttrl}
\begin{tabularx}{\columnwidth}{>{\raggedright\arraybackslash}p{4.4cm}@{\hspace{3pt}}*{5}{>{\centering\arraybackslash}X}@{\hspace{2pt}\color{black!18}\vrule width 0.5pt\hspace{4pt}}>{\columncolor{ExpHeader!70}\centering\arraybackslash}X}
\toprule
\rowcolor{ExpHeader}
\makecell[l]{\rule{0pt}{2.2ex}\textbf{Configuration}} & \benchsingle{NL4OPT} & \benchcell{MAMO}{Easy} & \benchcell{MAMO}{Complex} & \benchcell{Industry}{OR} & \benchsingle{OptMATH} & \benchcell{Macro}{Avg.} \\
\midrule
\multicolumn{7}{l}{\textit{Zero-shot}} \\
\hspace{1.1em}Qwen3-4B-Instruct-2507 & 55.9 & 81.9 & 11.8 & 27.0 & 10.8 & 37.5 \\
\addlinespace[2pt]
\multicolumn{7}{l}{\textit{Search Only (No Adaptation)}} \\
\hspace{1.1em}AutoFormulator (MCTS) & 75.5 & 85.7 & 23.2 & 30.0 & 11.4 & 45.2  \\
\hspace{1.1em}StarOR w/o LoRA& 89.4 & 90.0 & 49.3 & 47.0 & 28.3 & 60.8 \\
\addlinespace[2pt]
\multicolumn{7}{l}{\textit{Search + Adaptation (Ours)}} \\
\rowcolor{ExpAccent}
\hspace{1.1em}\textbf{StarOR w/ LoRA} & \textbf{92.7} & \textbf{94.4} & \textbf{54.2} & \textbf{51.0} & \textbf{32.5} & \textbf{65.0} \\
\bottomrule
\end{tabularx}
\end{table}

\paragraph{Impact of Reward Design (RQ3).}
Table~\ref{tab:rq3-reward} evaluates the effect of reward design in test-time adaptation. The multi-faceted reward achieves the best average accuracy, which underscores that simple consensus is insufficient for complex OR modeling; instead, rich and multi-dimensional feedback is essential to guide policy evolution. Table~\ref{tab:rq3-2reward} further shows that reward components are especially useful on complex benchmarks: removing them consistently degrades OptMATH performance. On the easier NL4OPT benchmark, however, some components yield neutral or negative effects, mainly because valid consensus is easier to obtain and imperfect pre-generation priors may introduce additional noise. We provide a detailed analysis of the pre-generation in the appendix~\ref{app:more}.

\begin{table*}[htbp]
\centering
\footnotesize
\setlength{\tabcolsep}{3.4pt}
\renewcommand{\arraystretch}{1.12}
\vspace{-0.3em}
\caption{(RQ3-1) Impact of reward design. We compare a simple majority-voting reward against the full multi-faceted reward under StarOR with Acc (\%).}
\vspace{-1.0em}
\label{tab:rq3-reward}
\begin{tabularx}{\textwidth}{>{\raggedright\arraybackslash}p{4.55cm}@{\hspace{3pt}}*{5}{>{\centering\arraybackslash}X}@{\hspace{2pt}\color{black!18}\vrule width 0.5pt\hspace{4pt}}>{\columncolor{ExpHeader!70}\centering\arraybackslash}X}
\toprule
\rowcolor{ExpHeader}
\makecell[l]{\rule{0pt}{2.2ex}\textbf{Method / Variant}} & \benchsingle{NL4OPT} & \benchcell{MAMO}{Easy} & \benchcell{MAMO}{Complex} & \benchcell{Industry}{OR} & \benchsingle{OptMATH} & \benchcell{Macro}{Avg.} \\
\midrule
\rowcolor{ExpBand}
\textbf{OR-R1-Qwen3-8B} & 88.3 & 86.1 & 49.9 & 35.3 & -- & -- \\
\multicolumn{7}{l}{\textbf{StarOR}} \\
\hspace{1.1em}+ No RL (Search Only) & 89.4 & 90.0 & 49.3 & 47.0 & 28.3 & 60.8 \\
\hspace{1.1em}+ RL (Major-voting Reward) & 90.6 & 92.4 & 48.3 & 46.0 & 26.5 & 60.8 \\
\rowcolor{ExpAccent}
\hspace{1.1em}+ RL (Multi-faceted Reward) & \textbf{92.7} & \textbf{94.4} & \textbf{54.2} & \textbf{51.0} & \textbf{32.5} & \textbf{65.0} \\
\bottomrule
\end{tabularx}
\end{table*}

\begin{table}[H]
\centering
\footnotesize
\vspace{-1.5em}
\caption{(RQ3-2) Detailed analysis for reward ablation on OptMATH (Complex) and NL4OPT (Easy). $\Delta$ reports the change relative to the full StarOR reward on each dataset.}
\label{tab:rq3-2reward}
\setlength{\tabcolsep}{8pt}
\renewcommand{\arraystretch}{1.15}
\begin{tabular}{l cc >{\hspace{10pt}}c c}
\toprule
\rowcolor{ExpHeader}
\textbf{Variant} 
& \multicolumn{2}{c}{\textbf{OptMATH}} 
& \multicolumn{2}{>{\hspace{10pt}}c}{\textbf{NL4OPT}} \\
\cmidrule(lr){2-3} \cmidrule(lr){4-5}
\rowcolor{ExpHeader}
& \textbf{Acc. (\%)} & \textbf{$\Delta$}
& \textbf{Acc. (\%)} & \textbf{$\Delta$} \\
\midrule
\rowcolor{ExpAccent}
\textbf{StarOR (Ours)} 
& \textbf{32.5} & -- 
& \textbf{92.7} & -- \\
\midrule
w/o objective scale 
& 29.5 & \textbf{-3.0}
& 93.1 & \textbf{+0.4} \\
\rowcolor{ExpBand}
w/o test-case reward ($r_{\mathrm{test}}$) 
& 30.7 & \textbf{-1.8}
& 92.2 & \textbf{-0.5} \\
w/o dynamic shaping 
& 31.3 & \textbf{-1.2}
& 92.7 & \textbf{+0.0} \\
\bottomrule
\end{tabular}
\end{table}

\paragraph{Test-Time Scaling Analysis (RQ4).} Table~\ref{tab:scaling} illustrates how OptMATH performance scales with the expansion group size $K$. We observe a consistent upward trend as $K$ increases from 2 to 32, demonstrating StarOR's ability to leverage additional test-time compute. StarOR with only $K=4$ reaches 25.9\% accuracy in 93.9s, compared with 24.9\% accuracy in 104.2s for Best-of-$N$ with $N=32$. This scaling behavior validates its effectiveness for complex scenarios where precision benefits from extended search. A detailed analysis for cost and efficiency is in Appendix~\ref{app:cost}.

\begin{table*}[htbp]
\centering
\footnotesize
\setlength{\tabcolsep}{2pt} 
\renewcommand{\arraystretch}{1.12}
\vspace{-0.3em}
\caption{(RQ4) We compare StarOR's expansion-based scaling against Best-of-$N$ on OptMATH. StarOR results are single-seed accuracies; Best-of-$N$ results are three-seed means. $\Delta$ is the difference between the displayed accuracies and the corresponding $K=2$ or $N=2$ result.}
\vspace{-1.0em}
\label{tab:scaling}
\begin{tabularx}{\textwidth}{
    >{\hsize=0.8\hsize\centering\arraybackslash}X 
    >{\hsize=0.7\hsize\centering\arraybackslash}X 
    >{\hsize=0.6\hsize\centering\arraybackslash}X 
    >{\hsize=0.9\hsize\centering\arraybackslash}X 
    @{\hspace{4pt}\color{black!18}\vrule width 0.5pt\hspace{6pt}}
    >{\hsize=0.8\hsize\centering\arraybackslash}X
    >{\hsize=0.7\hsize\centering\arraybackslash}X
    >{\hsize=0.6\hsize\centering\arraybackslash}X
    >{\hsize=0.9\hsize\centering\arraybackslash}X 
}
\toprule
\rowcolor{ExpHeader}
\textbf{Num $K$} & \textbf{Acc (\%)} & \textbf{$\Delta$(\%)} & \textbf{Time Cost (s)} & \textbf{Best-of-N} & \textbf{Acc (\%)} & \textbf{$\Delta$(\%)} & \textbf{Time Cost (s)} \\
\midrule
$K=2$  & 23.5 & --    & 56.1 & $N=2$  & 12.5 & -- & 37.2 \\
\rowcolor{ExpBand}
$K=4$  & 25.9 & \textbf{+2.4}  & 93.9 & $N=4$  & 15.9 & \textbf{+3.4} & 46.4 \\
$K=8$  & 32.5 & \textbf{+9.0}  & 205.2 & $N=8$  & 20.3 & \textbf{+7.8} & 62.9 \\
\rowcolor{ExpBand}
$K=16$ & 34.9 & \textbf{+11.4} & 575.7 & $N=16$ & 21.9 & \textbf{+9.4} & 81.6 \\

$K=32$ & 37.3 & \textbf{+13.8} & 1396.1 & $N=32$ & 24.9 & \textbf{+12.4} & 104.2 \\
\bottomrule
\end{tabularx}
\end{table*}
\vspace{-0.5em}

\section{Conclusion}
\label{conclusion}

In this paper, we revisit optimization modeling as a hierarchical decision-making process rather than a flat text-to-code generation task. Due to the brittleness of hierarchical formulation, one-shot generation is prone to error propagation, while fixed-policy search may repeatedly explore flawed modeling paths. To address this challenge, we proposed \textsc{StarOR}, a framework integrating stage-wise MCTS with node-level test-time reinforcement learning. By decomposing formulation into discrete stages and updating a transient LoRA adapter via GRPO, \textsc{StarOR} transforms execution-driven feedback into instance-specific policy refinement, with an unsupervised multi-faceted reward providing fine-grained node-level supervision. Across five benchmarks, \textsc{StarOR} reaches the highest reported average accuracy with a 4B backbone, demonstrating the effectiveness of search-driven test-time evolution for optimization modeling.

\newpage

\bibliographystyle{iclr2027_conference}
\bibliography{ref}

@article{huang2024orlm,
  title = {ORLM: A Customizable Framework in Training Large Models for Automated Optimization Modeling},
  author = {Huang, Chenyu and Tang, Zhengyang and Hu, Shixi and Jiang, Ruoqing and Zheng, Xin and Ge, Dongdong and Wang, Benyou and Wang, Zizhuo},
  journal = {arXiv preprint arXiv:2405.17743},
  year = {2024},
  url = {https://arxiv.org/abs/2405.17743}
}

@article{jiang2024llmopt,
  title = {LLMOPT: Learning to Define and Solve General Optimization Problems from Scratch},
  author = {Jiang, Caigao and Shu, Xiang and Qian, Hong and Lu, Xingyu and Zhou, Jun and Zhou, Aimin and Yu, Yang},
  journal = {arXiv preprint arXiv:2410.13213},
  year = {2024},
  url = {https://arxiv.org/abs/2410.13213}
}

@article{lu2025optmath,
  title = {OptMATH: A Scalable Bidirectional Data Synthesis Framework for Optimization Modeling},
  author = {Lu, Hongliang and Xie, Zhonglin and Wu, Yaoyu and Ren, Can and Chen, Yuxuan and Wen, Zaiwen},
  journal = {arXiv preprint arXiv:2502.11102},
  year = {2025},
  url = {https://arxiv.org/abs/2502.11102}
}

@article{chen2025sirl,
  title = {Solver-Informed RL: Grounding Large Language Models for Authentic Optimization Modeling},
  author = {Chen, Yitian and Xia, Jingfan and Shao, Siyu and Ge, Dongdong and Ye, Yinyu},
  journal = {arXiv preprint arXiv:2505.11792},
  year = {2025},
  url = {https://arxiv.org/abs/2505.11792}
}

@article{solverllm2025,
  title = {SolverLLM: Leveraging Test-Time Scaling for Optimization Problem via LLM-Guided Search},
  author = {Li, Dong and Zhao, Xujiang and Yu, Linlin and Liu, Yanchi and Cheng, Wei and Chen, Zhengzhang and Chen, Zhong and Chen, Feng and Zhao, Chen and Chen, Haifeng},
  journal = {arXiv preprint arXiv:2510.16916},
  year = {2025},
  url = {https://arxiv.org/abs/2510.16916}
}

@inproceedings{astorga2025autoformulation,
  title = {Autoformulation of Mathematical Optimization Models Using LLMs},
  author = {Astorga, Nicol{\'a}s and Liu, Tennison and Xiao, Yuanzhang and Van Der Schaar, Mihaela},
  booktitle = {Proceedings of the 42nd International Conference on Machine Learning},
  pages = {1864--1886},
  year = {2025},
  volume = {267},
  series = {Proceedings of Machine Learning Research},
  url = {https://proceedings.mlr.press/v267/astorga25a.html}
}

@inproceedings{liu2025optitree,
  title = {OptiTree: Hierarchical Thoughts Generation with Tree Search for LLM Optimization Modeling},
  author = {Liu, Haoyang and Wang, Jie and Cai, Yuyang and Han, Xiongwei and Kuang, Yufei and Hao, Jianye},
  booktitle = {Advances in Neural Information Processing Systems},
  year = {2025},
  url = {https://openreview.net/forum?id=Ej20yjWMCj}
}

@inproceedings{zuo2025ttrl,
  title = {TTRL: Test-Time Reinforcement Learning},
  author = {Zuo, Yuxin and Zhang, Kaiyan and Sheng, Li and Qu, Shang and Cui, Ganqu and Zhu, Xuekai and Li, Haozhan and Zhang, Yuchen and Long, Xinwei and Hua, Ermo and Qi, Biqing and Sun, Youbang and Ma, Zhiyuan and Yuan, Lifan and Ding, Ning and Zhou, Bowen},
  booktitle = {Advances in Neural Information Processing Systems},
  year = {2025},
  url = {https://openreview.net/forum?id=VuVhgEiu20}
}

@article{jiao2026policy,
  title = {Policy of Thoughts: Scaling LLM Reasoning via Test-time Policy Evolution},
  author = {Jiao, Zhengbo and Xian, Hongyu and Wang, Qinglong and Ma, Yunpu and Wang, Zhebo and Zhang, Zifan and Kong, Dezhang and Han, Meng},
  journal = {arXiv preprint arXiv:2601.20379},
  year = {2026},
  url = {https://arxiv.org/abs/2601.20379}
}

@article{ding2025orr1,
  title = {OR-R1: Automating Modeling and Solving of Operations Research Optimization Problem via Test-Time Reinforcement Learning},
  author = {Ding, Zezhen and Tan, Zhen and Zhang, Jiheng and Chen, Tianlong},
  journal = {arXiv preprint arXiv:2511.09092},
  year = {2025},
  url = {https://arxiv.org/abs/2511.09092}
}

@article{shao2024deepseekmath,
  title = {DeepSeekMath: Pushing the Limits of Mathematical Reasoning in Open Language Models},
  author = {Shao, Zhihong and Wang, Peiyi and Zhu, Qihao and Xu, Runxin and Song, Junxiao and Bi, Xiao and Zhang, Haowei and Zhang, Mingchuan and Li, Y. K. and Wu, Y. and Guo, Daya},
  journal = {arXiv preprint arXiv:2402.03300},
  year = {2024},
  url = {https://arxiv.org/abs/2402.03300}
}

@inproceedings{ramamonjison2022augmenting,
  title = {Augmenting Operations Research with Auto-Formulation of Optimization Models From Problem Descriptions},
  author = {Ramamonjison, Rindra and Li, Haley and Yu, Timothy and He, Shiqi and Rengan, Vishnu and Banitalebi-dehkordi, Amin and Zhou, Zirui and Zhang, Yong},
  booktitle = {Proceedings of the 2022 Conference on Empirical Methods in Natural Language Processing: Industry Track},
  pages = {29--62},
  year = {2022},
  address = {Abu Dhabi, UAE},
  publisher = {Association for Computational Linguistics},
  doi = {10.18653/v1/2022.emnlp-industry.4},
  url = {https://aclanthology.org/2022.emnlp-industry.4/}
}

@inproceedings{nl4opt2022,
  title = {NL4Opt Competition: Formulating Optimization Problems Based on Their Natural Language Descriptions},
  author = {Ramamonjison, Rindranirina and Yu, Timothy and Li, Raymond and Li, Haley and Carenini, Giuseppe and Ghaddar, Bissan and He, Shiqi and Mostajabdaveh, Mahdi and Banitalebi-Dehkordi, Amin and Zhou, Zirui and Zhang, Yong},
  booktitle = {Proceedings of the NeurIPS 2022 Competitions Track},
  pages = {189--203},
  year = {2022},
  volume = {220},
  series = {Proceedings of Machine Learning Research},
  publisher = {PMLR},
  url = {https://proceedings.mlr.press/v220/ramamonjison23a.html}
}

@article{ahmed2024lm4opt,
  title = {LM4OPT: Unveiling the Potential of Large Language Models in Formulating Mathematical Optimization Problems},
  author = {Ahmed, Tasnim and Choudhury, Salimur},
  journal = {INFOR: Information Systems and Operational Research},
  volume = {62},
  number = {4},
  pages = {559--572},
  year = {2024},
  doi = {10.1080/03155986.2024.2388452},
  url = {https://www.tandfonline.com/doi/abs/10.1080/03155986.2024.2388452}
}

@inproceedings{wu2025stepopt,
  title = {Training LLMs for Optimization Modeling via Iterative Data Synthesis and Structured Validation},
  author = {Wu, Yang and Zhang, Yifan and Wu, Yurong and Wang, Yuran and Zhang, Junkai and Cheng, Jian},
  booktitle = {Findings of the Association for Computational Linguistics: EMNLP 2025},
  pages = {12880--12896},
  year = {2025},
  address = {Suzhou, China},
  publisher = {Association for Computational Linguistics},
  doi = {10.18653/v1/2025.findings-emnlp.691},
  url = {https://aclanthology.org/2025.findings-emnlp.691/}
}

@inproceedings{wang2025ormind,
  title = {ORMind: A Cognitive-Inspired End-to-End Reasoning Framework for Operations Research},
  author = {Wang, Zhiyuan and Chen, Bokui and Huang, Yinya and Cao, Qingxing and He, Ming and Fan, Jianping and Liang, Xiaodan},
  booktitle = {Proceedings of the 63rd Annual Meeting of the Association for Computational Linguistics (Volume 6: Industry Track)},
  pages = {104--131},
  year = {2025},
  address = {Vienna, Austria},
  publisher = {Association for Computational Linguistics},
  doi = {10.18653/v1/2025.acl-industry.10},
  url = {https://aclanthology.org/2025.acl-industry.10/}
}

@misc{wang2025bpp,
      title={BPP-Search: Enhancing Tree of Thought Reasoning for Mathematical Modeling Problem Solving}, 
      author={Teng Wang and Wing-Yin Yu and Zhenqi He and Zehua Liu and Hailei Gong and Han Wu and Xiongwei Han and Wei Shi and Ruifeng She and Fangzhou Zhu and Tao Zhong},
      year={2025},
      eprint={2411.17404},
      archivePrefix={arXiv},
      primaryClass={cs.AI},
      url={https://arxiv.org/abs/2411.17404}, 
}

@inproceedings{wang2023selfconsistency,
  title = {Self-Consistency Improves Chain of Thought Reasoning in Language Models},
  author = {Wang, Xuezhi and Wei, Jason and Schuurmans, Dale and Le, Quoc V. and Chi, Ed H. and Narang, Sharan and Chowdhery, Aakanksha and Zhou, Denny},
  booktitle = {The Eleventh International Conference on Learning Representations},
  year = {2023},
  url = {https://openreview.net/forum?id=1PL1NIMMrw}
}

@inproceedings{yao2023tree,
  title = {Tree of Thoughts: Deliberate Problem Solving with Large Language Models},
  author = {Yao, Shunyu and Yu, Dian and Zhao, Jeffrey and Shafran, Izhak and Griffiths, Thomas L. and Cao, Yuan and Narasimhan, Karthik R.},
  booktitle = {Advances in Neural Information Processing Systems},
  year = {2023},
  url = {https://openreview.net/forum?id=5Xc1ecxO1h}
}

@misc{huang2024mamo,
      title={Mamo: a Mathematical Modeling Benchmark with Solvers}, 
      author={Xuhan Huang and Qingning Shen and Yan Hu and Anningzhe Gao and Benyou Wang},
      year={2024},
      eprint={2405.13144},
      archivePrefix={arXiv},
      primaryClass={cs.AI}
}

@InProceedings{xiao2024chainofexperts,
  title     = {Chain-of-Experts: When LLMs Meet Complex Operations Research Problems},
  author    = {Xiao, Ziyang and Zhang, Dongxiang and Wu, Yangjun and Xu, Lilin and Wang, Yuan Jessica and Han, Xiongwei and Fu, Xiaojin and Zhong, Tao and Zeng, Jia and Song, Mingli and Chen, Gang},
  booktitle = {The Twelfth International Conference on Learning Representations (ICLR)},
  year      = {2024},
  url       = {https://openreview.net/forum?id=HobyL1B9CZ}
}

@misc{gurobi,
  author = {{Gurobi Optimization, LLC}},
  title = {{Gurobi Optimizer Reference Manual}},
  year = {2026},
  url = {https://www.gurobi.com}
}

@article{achiam2023gpt4,
  title     = {GPT-4 Technical Report},
  author    = {OpenAI and Achiam, Josh and Adler, Steven and Agarwal, Sandhini and Ahmad, Lama and Akkaya, Ilge and Aleman, Florencia Leoni and Almeida, Diogo and others},
  journal   = {arXiv preprint arXiv:2303.08774},
  year      = {2023},
  eprint    = {2303.08774},
  archivePrefix = {arXiv},
  primaryClass  = {cs.CL},
  url       = {https://arxiv.org/abs/2303.08774},
  doi       = {10.48550/arXiv.2303.08774}
}

@article{guo2025deepseekr1,
  title     = {DeepSeek-R1: Incentivizing Reasoning Capability in LLMs via Reinforcement Learning},
  author    = {DeepSeek-AI and Guo, Daya and Yang, Dejian and Zhang, Haowei and Song, Junxiao and Wang, Peiyi and Zhu, Qihao and Xu, Runxin and Zhang, Ruoyu and Ma, Shirong and others},
  journal   = {Nature},
  volume    = {645},
  pages     = {633--638},
  year      = {2025},
  publisher = {Springer Nature},
  doi       = {10.1038/s41586-025-09422-z},
  note      = {Also available as arXiv preprint arXiv:2501.12948},
  url       = {https://arxiv.org/abs/2501.12948}
}

@article{deepseekai2024deepseekv3,
  title     = {DeepSeek-V3 Technical Report},
  author    = {DeepSeek-AI and Liu, Aixin and Feng, Bei and Xue, Bing and Wang, Bingxuan and Wu, Bochao and Lu, Chengda and Zhao, Chenggang and Deng, Chengqi and Zhang, Chenyu and others},
  journal   = {arXiv preprint arXiv:2412.19437},
  year      = {2024},
  eprint    = {2412.19437},
  archivePrefix = {arXiv},
  primaryClass  = {cs.CL},
  url       = {https://arxiv.org/abs/2412.19437},
  doi       = {10.48550/arXiv.2412.19437},
  note      = {DeepSeek-V3.1 is an incremental update based on this architecture}
}

@inproceedings{kang2025scalable,
  title     = {Scalable Best-of-N Selection for Large Language Models via Self-Certainty},
  author    = {Kang, Zhewei and Zhao, Xuandong and Song, Dawn},
  booktitle = {Advances in Neural Information Processing Systems (NeurIPS)},
  year      = {2025},
  volume    = {38},
  pages     = {},
  publisher = {Curran Associates, Inc.},
  url       = {https://arxiv.org/abs/2502.18581},
  eprint    = {2502.18581},
  archivePrefix = {arXiv},
  primaryClass  = {cs.CL}
}

@book{Belegundu_Chandrupatla_2019, place={Cambridge}, edition={3}, title={Optimization Concepts and Applications in Engineering}, publisher={Cambridge University Press}, author={Belegundu, Ashok D. and Chandrupatla, Tirupathi R.}, year={2019}}

@article{singh2012overview,
  author    = {Singh, Ajay},
  title     = {An overview of the optimization modelling applications},
  journal   = {Journal of Hydrology},
  year      = {2012},
  volume    = {466--467},
  pages     = {167--182},
  month     = oct,
  doi       = {10.1016/j.jhydrol.2012.08.004},
  issn      = {0022-1694},
  publisher = {Elsevier},
  type      = {Review Paper}
}

@article{krishnamurthy2018energy,
  author    = {Krishnamurthy, Dheepak and Uckun, Canan and Zhou, Zhi and Thimmapuram, Prakash R. and Botterud, Audun},
  title     = {Energy Storage Arbitrage Under Day-Ahead and Real-Time Price Uncertainty},
  journal   = {IEEE Transactions on Power Systems},
  year      = {2018},
  volume    = {33},
  number    = {1},
  pages     = {84--93},
  doi       = {10.1109/TPWRS.2017.2685347},
  issn      = {0885-8950},
  publisher = {IEEE}
}

@inproceedings{shinn2023reflexion,
  title     = {Reflexion: Language Agents with Verbal Reinforcement Learning},
  author    = {Shinn, Noah and Cassano, Federico and Berman, Edward and Gopinath, Ashwin and Narasimhan, Karthik and Yao, Shunyu},
  booktitle = {Advances in Neural Information Processing Systems (NeurIPS)},
  year      = {2023},
  volume    = {36},
  pages     = {39870--39890},
  publisher = {Curran Associates, Inc.},
  url       = {https://arxiv.org/abs/2303.11366},
  eprint    = {2303.11366},
  archivePrefix = {arXiv},
  primaryClass  = {cs.AI}
}

@misc{lian2026reloop,
      title={ReLoop: Structured Modeling and Behavioral Verification for Reliable LLM-Based Optimization}, 
      author={Junbo Jacob Lian and Yujun Sun and Huiling Chen and Chaoyu Zhang and Hanzhang Qin and Chung-Piaw Teo},
      year={2026},
      eprint={2602.15983},
      archivePrefix={arXiv},
      primaryClass={cs.SE},
      url={https://arxiv.org/abs/2602.15983}, 
}

@misc{zhang2025sacopt,
  title         = {SAC-Opt: Semantic Anchors for Iterative Correction in Optimization Modeling},
  author        = {Yansen Zhang and Qingcan Kang and Yujie Chen and Yufei Wang and Xiongwei Han and Tao Zhong and Mingxuan Yuan and Chen Ma},
  year          = {2025},
  eprint        = {2510.05115},
  archivePrefix = {arXiv},
  primaryClass  = {cs.AI},
  url           = {https://arxiv.org/abs/2510.05115}
}

@inproceedings{akyurek2025surprising,
  title     = {The Surprising Effectiveness of Test-Time Training for Few-Shot Learning},
  author    = {Aky{\"u}rek, Ekin and Damani, Mehul and Zweiger, Adam and Qiu, Linlu and Guo, Han and Pari, Jyothish and Kim, Yoon and Andreas, Jacob},
  booktitle = {Proceedings of the 42nd International Conference on Machine Learning},
  year      = {2025},
  url       = {https://openreview.net/forum?id=asgBo3FNdg}
}

@article{liao2026tool,
  title   = {Tool Verification for Test-Time Reinforcement Learning},
  author  = {Liao, Ruotong and R{\"o}hrich, Nikolai and Wang, Xiaohan and Zhang, Yuhui and Samadzadeh, Yasaman and Tresp, Volker and Yeung-Levy, Serena},
  journal = {arXiv preprint arXiv:2603.02203},
  year    = {2026},
  url     = {https://arxiv.org/abs/2603.02203}
}

@article{wang2025thetaevolve,
  title   = {{ThetaEvolve}: Test-time Learning on Open Problems},
  author  = {Wang, Yiping and Su, Shao-Rong and Zeng, Zhiyuan and Xu, Eva and Ren, Liliang and Yang, Xinyu and Huang, Zeyi and He, Xuehai and Ma, Luyao and Peng, Baolin and Cheng, Hao and He, Pengcheng and Chen, Weizhu and Wang, Shuohang and Du, Simon Shaolei and Shen, Yelong},
  journal = {arXiv preprint arXiv:2511.23473},
  year    = {2025},
  url     = {https://arxiv.org/abs/2511.23473}
}

@article{yuksekgonul2026learning,
  title   = {Learning to Discover at Test Time},
  author  = {Yuksekgonul, Mert and Koceja, Daniel and Li, Xinhao and Bianchi, Federico and McCaleb, Jed and Wang, Xiaolong and Kautz, Jan and Choi, Yejin and Zou, James and Guestrin, Carlos and Sun, Yu},
  journal = {arXiv preprint arXiv:2601.16175},
  year    = {2026},
  url     = {https://arxiv.org/abs/2601.16175}
}

@misc{hu2025tlm,
      title={Test-Time Learning for Large Language Models}, 
      author={Jinwu Hu and Zhitian Zhang and Guohao Chen and Xutao Wen and Chao Shuai and Wei Luo and Bin Xiao and Yuanqing Li and Mingkui Tan},
      year={2025},
      eprint={2505.20633},
      archivePrefix={arXiv},
      primaryClass={cs.CL},
      url={https://arxiv.org/abs/2505.20633}, 
}

@misc{openai2025o3,
  title        = {Introducing OpenAI o3 and o4-mini},
  author       = {{OpenAI}},
  year         = {2025},
  month        = apr,
  day          = {16},
  howpublished = {\url{https://openai.com/index/introducing-o3-and-o4-mini/}},
  note         = {Accessed: 2026-05-05}
}

@article{sheng2024hybridflow,
  title   = {HybridFlow: A Flexible and Efficient RLHF Framework},
  author  = {Guangming Sheng and Chi Zhang and Zilingfeng Ye and Xibin Wu and Wang Zhang and Ru Zhang and Yanghua Peng and Haibin Lin and Chuan Wu},
  year    = {2024},
  journal = {arXiv preprint arXiv: 2409.19256}
}

\newpage
\appendix

\definecolor{PromptTitleBlue}{RGB}{92,137,174}
\definecolor{PromptBorderBlue}{RGB}{86,132,170}
\definecolor{PromptBodyBlue}{RGB}{238,244,249}

\newtcblisting{starorprompt}[2][]{%
  enhanced,
  breakable,
  listing only,
  colback=PromptBodyBlue,
  colframe=PromptBorderBlue,
  colbacktitle=PromptTitleBlue,
  coltitle=white,
  title={#2},
  fonttitle=\large\bfseries\rmfamily,
  lefttitle=4mm,
  righttitle=4mm,
  toptitle=1.2mm,
  bottomtitle=1.2mm,
  arc=4pt,
  outer arc=4pt,
  boxrule=0.9pt,
  left=4mm,
  right=4mm,
  top=3.8mm,
  bottom=3mm,
  before skip=8pt,
  after skip=8pt,
  listing options={
    basicstyle=\small\rmfamily,
    breaklines=true,
    columns=fullflexible,
    keepspaces=true,
    showstringspaces=false,
    tabsize=2
  },
  #1
}

\section{Discussion}

\paragraph{AI Usage Statement}
Generative AI tools assisted with research methodology and experiment design, method implementation, and interpretation of experimental results. They also assisted with drafting, translating, and editing manuscript text; literature and reference work; and preparing or revising figures. The authors reviewed AI-assisted text, translations, figures, citations, and result interpretations against the underlying sources and data, and ran and tested AI-assisted code. The authors take responsibility for the final content of this work, including text, claims, code, and figures produced with AI assistance.

\paragraph{limitation}
\label{app:limitations}

StarOR incurs higher test-time cost than one-shot decoding and Best-of-$N$ with the same number of raw samples, since it performs staged search, repeated code execution, reward computation, and transient LoRA updates. However, in real industrial optimization scenarios, especially high-complexity and high-value applications, reliability is often more important than real-time response. For many offline optimization-modeling settings, this additional cost is therefore acceptable when it leads to more faithful and robust formulations.

Another limitation is that StarOR relies on unsupervised reward signals, particularly objective-scale priors and synthetic robustness tests. These signals are useful for guiding search without ground-truth labels, but their quality depends on the reasoning ability of the backbone model. For weaker small models, inaccurate objective-scale estimation or flawed perturbation analysis may introduce noisy feedback and mislead policy adaptation. This suggests that StarOR still requires a reasonably capable base model. Future work should design stronger reward and feedback mechanisms that provide more fine-grained and reliable supervision, thereby better guiding both policy evolution and tree search during test-time optimization.

\section{Baselines}
\label{app:baselines}
\label{appendix:baselines}

We compare StarOR with three families of baselines: strong general-purpose LLMs, offline learning-based OR modeling systems, and inference-time scaling methods. For all test-time baselines implemented in our codebase, we use the same backbone, Qwen3-4B-Instruct-2507, and the same Gurobi execution environment as StarOR unless otherwise stated. This controls for differences in base model capability and isolates the effect of the test-time algorithm.

\paragraph{Zero-shot.}
The zero-shot baseline directly prompts the backbone model to generate the complete optimization formulation and solver-ready Gurobi Python program in a single pass. The output is executed once and the resulting objective value is compared against the reference answer. This baseline measures the raw modeling ability of the backbone without sampling, search, repair, or adaptation.

\paragraph{Best-of-N.}
Best-of-N is a standard inference-time scaling baseline inspired by self-consistency and repeated sampling \citep{wang2023selfconsistency,kang2025scalable}. We sample $N$ independent solver-code candidates from the same backbone and execute each candidate with a $30$ second solver/execution budget. The final answer is selected from executable candidates by objective-value consensus: candidates are clustered by numerical objective value using a relative tolerance of $10^{-4}$, and the largest executable cluster is selected; ties are resolved by execution validity and generation order. In the main comparison, we use $N=16$ with temperature $1.0$ and maximum generation length $8196$ tokens. In the scaling study, $N$ is swept over $\{2,4,8,16,32\}$ to match the compute axis used by StarOR.

\paragraph{Reflexion.}
Reflexion \citep{shinn2023reflexion} is implemented as an iterative generate-execute-refine loop. A candidate Gurobi program is first generated and executed. The execution trace, including Python exceptions, solver status, infeasibility/unboundedness messages, or suspicious missing objective output, is then fed back to the same model as verbal feedback for code refinement. We run up to $10$ refinement attempts with maximum generation length $8196$ tokens and a $30$ second execution budget per attempt. The final candidate is the first executable candidate that passes the solver checks; if no such candidate exists, we select the highest-quality candidate according to the same execution and objective-consensus rule used for Best-of-$N$.

\paragraph{AutoFormulator.}
AutoFormulator \citep{astorga2025autoformulation} also uses hierarchical MCTS to assemble optimization formulations. It ranks partial formulations with an LLM, prunes equivalent candidates, and backs up LLM-and-solver rewards from complete leaf formulations; its generator remains fixed during inference. The central distinction is that StarOR evaluates partial modeling decisions through executable completions at each search expansion and uses the resulting rewards both to guide MCTS and to update a transient, instance-specific LoRA policy with GRPO. Search therefore changes StarOR's subsequent proposal distribution, rather than only the branch-selection statistics.

\paragraph{OptiTree.}
OptiTree \citep{liu2025optitree} constructs a reusable modeling tree from a problem dataset, with nodes storing problem categories and high-level modeling thoughts. At inference, it searches that tree for relevant subproblems and retrieves their thoughts to guide formulation. The central distinction is the object of search: OptiTree retrieves from a preconstructed knowledge tree, whereas StarOR searches the current instance's evolving partial formulation. StarOR further evaluates those decisions through executable rollouts and updates its policy during the same search, coupling formulation exploration with instance-specific learning.

\paragraph{OR-R1.}
OR-R1 \citep{ding2025orr1} is a learning-based OR modeling system that combines supervised fine-tuning with test-time group relative policy optimization. Its reward contains OR-specific components such as format validity, code validity, code executability, solution correctness, and consistency. We include the reported OR-R1-Qwen3-8B results as a strong TTRL-oriented baseline. The comparison is intentionally conservative: OR-R1 uses a larger 8B backbone and task-specific training (SFT), whereas StarOR uses a 4B backbone and performs instance-level adaptation with a transient LoRA adapter during search.

\paragraph{Learning-based baselines.}
We also report specialized offline-training methods, including ORLM \citep{huang2024orlm}, LLMOPT \citep{jiang2024llmopt}, OptMATH \citep{lu2025optmath}, and SIRL \citep{chen2025sirl}. These methods improve LLMs through synthetic data, solver-informed supervision, or reinforcement learning before evaluation. They provide a reference for how far static training can push OR modeling accuracy, while StarOR studies the complementary direction of spending compute at test time on each instance. The ORLM row in Table~\ref{tab:main-results} follows the five-benchmark comparison in \citet{chen2025sirl}, including its 24.0\% IndustryOR result.

\paragraph{TTT-Discover.}
TTT-Discover \citep{yuksekgonul2026learning} combines PUCT-guided reuse of evaluated solutions with instance-level test-time reinforcement learning under a directly measurable continuous reward, such as inverse kernel runtime. In OR modeling, a candidate's solver objective alone cannot establish whether its formulation faithfully represents the problem, and the reference optimum is unavailable at test time. To our knowledge, StarOR is the first framework to combine hierarchical MCTS with test-time training for OR modeling. Its key contribution is a tailored unsupervised, multi-faceted reward: objective-value consensus, executability, test-case robustness, and structural consistency jointly guide search and learning without a ground-truth correctness score. The design motivation also differs. TTT-Discover seeks to improve candidate solutions against a directly measurable reward, whereas StarOR addresses the fragility of interdependent formulation decisions, where an early error in sets or variables can compromise downstream constraints. StarOR therefore searches four stages of partial modeling decisions, completes each decision to executable code for evaluation, and uses the resulting stage-level feedback for both MCTS and transient policy updates. In high-value OR modeling, this emphasis on formulation fidelity can justify additional test-time computation.

\section{More Experiment Results and Analysis}
\label{app:more}

Unless otherwise stated, all reward weights are ordered as
$(w_{\mathrm{sem}}, w_{\mathrm{exec}}, w_{\mathrm{test}}, w_{\mathrm{struct}})$,
corresponding to semantic consensus, executability, test-case robustness, and structural consistency.  In the full StarOR configuration, we use dynamic reward shaping: the weights are $(0.2,0.5,0.2,0.1)$ for iterations 1--3, $(0.4,0.4,0.1,0.1)$ for iterations 4--5, and $(0.6,0.2,0.1,0.1)$ afterwards.  For the \emph{w/o dynamic shaping} ablation in Table~\ref{tab:rq3-2reward}, we disable this schedule and use the fixed late-stage weight vector $(0.6,0.2,0.1,0.1)$ throughout all MCTS iterations.  For the \emph{w/o test-case reward} ablation, we remove $r_{\mathrm{test}}$ from the reward computation and transfer its weight to $r_{\mathrm{exec}}$ so that the total reward scale remains unchanged.

\begin{table}[H]
\centering
\footnotesize
\caption{Reward-weight settings for the reward ablations. Tuples are ordered as $(w_{\mathrm{sem}}, w_{\mathrm{exec}}, w_{\mathrm{test}}, w_{\mathrm{struct}})$.}
\label{tab:appendix-reward-ablation-settings}
\setlength{\tabcolsep}{5pt}
\renewcommand{\arraystretch}{1.12}
\begin{tabular}{l c c c}
\toprule
\rowcolor{ExpHeader}
\textbf{Setting} & \makecell{\textbf{Iter.}\\\textbf{1--3}} & \makecell{\textbf{Iter.}\\\textbf{4--5}} & \makecell{\textbf{Iter.}\\\textbf{$>5$}} \\
\midrule
Full StarOR & $(0.2,0.5,0.2,0.1)$ & $(0.4,0.4,0.1,0.1)$ & $(0.6,0.2,0.1,0.1)$ \\
\rowcolor{ExpBand}
w/o dynamic shaping & $(0.6,0.2,0.1,0.1)$ & $(0.6,0.2,0.1,0.1)$ & $(0.6,0.2,0.1,0.1)$ \\
w/o test-case reward & $(0.2,0.7,0.0,0.1)$ & $(0.4,0.5,0.0,0.1)$ & $(0.6,0.3,0.0,0.1)$ \\
\bottomrule
\end{tabular}
\end{table}

\subsection{Best-of-N Baseline Definition}
For the Best-of-$N$ scaling comparison in Table~\ref{tab:scaling}, we sample $N$ independent code candidates in one batch and execute their programs in parallel. We select the answer by objective-value consensus among executable candidates, as described in Appendix~\ref{app:baselines}. Wall time is measured end to end, including candidate generation and code execution. To account for sampling variability, we repeat each setting with three seeds. Table~\ref{tab:bon-multiseed} reports mean accuracy and sample standard deviation, together with the observed average completion tokens and wall time. These $N$ settings measure scaling with sample count; the separate budget-controlled comparison is reported in Section~\ref{app:compute-budget}.

\begin{table}[H]
\centering
\footnotesize
\caption{Three-seed Best-of-$N$ results on OptMATH. Accuracy is mean $\pm$ sample standard deviation; completion tokens and end-to-end wall time are observed averages per instance.}
\label{tab:bon-multiseed}
\setlength{\tabcolsep}{8pt}
\renewcommand{\arraystretch}{1.12}
\begin{tabular}{c c c c}
\toprule
\rowcolor{ExpHeader}
\textbf{Best-of-$N$} & \textbf{Acc. (\%)} & \makecell{\textbf{Completion}\\\textbf{tokens (K)}} & \makecell{\textbf{Wall time}\\\textbf{(s)}} \\
\midrule
$N=2$ & $12.45\pm0.70$ & 8.6 & 37.23 \\
\rowcolor{ExpBand}
$N=4$ & $15.86\pm0.92$ & 17.4 & 46.36 \\
$N=8$ & $20.28\pm1.94$ & 36.1 & 62.91 \\
\rowcolor{ExpBand}
$N=16$ & $21.89\pm0.35$ & 71.3 & 81.55 \\
$N=32$ & $24.90\pm1.94$ & 141.8 & 104.21 \\
\bottomrule
\end{tabular}
\end{table}

Accuracy increases with $N$, alongside higher token use and wall time. Because these rows vary the sample count, they should be interpreted as a scaling study rather than an equal-budget comparison.

\subsection{Pre-generation analysis}
Before running the stage-wise search, StarOR constructs a reward prior by estimating a conservative objective-scale envelope $[L_x,U_x]$ for each instance. This pre-generation signal is not used as a hard correctness oracle; rather, it provides a soft preference that down-weights candidates whose objective values fall outside the estimated range. Ground-truth objectives are used only for the retrospective diagnostics in Table~\ref{tab:pregeneration-analysis}, never to generate the prior or guide test-time search.

We report four diagnostic statistics. First, \emph{GT covered by base scale} measures the percentage of instances whose ground-truth objective lies within the estimated range $[L_x,U_x]$, reflecting the coverage quality of the pre-generated scale. Second, we report the conditional accuracy when the ground-truth objective is inside the estimated range, which measures how well StarOR performs when the prior is consistent with the true objective scale. Third, we report the conditional accuracy when the ground-truth objective falls outside the estimated range, which evaluates whether StarOR can still recover the correct answer when the prior is inaccurate. Finally, $\Delta$ Acc. measures the accuracy gap between these two subsets, quantifying the performance drop caused by an incorrect objective-scale estimate.

Overall, the results show that the estimated base scale covers the ground-truth objective for a large fraction of instances across datasets, while StarOR still retains non-trivial recovery ability when the ground truth falls outside the predicted range. The trends also reveal clear differences in dataset difficulty. On relatively easier datasets such as NL4OPT and MAMO-Easy, StarOR achieves high conditional accuracy when the ground truth is covered by the base scale, and MAMO-Easy in particular remains robust even when the scale estimate is incorrect. In contrast, harder datasets such as MAMO-Complex, IndustryOR, and especially OptMATH exhibit substantially lower conditional accuracies in both subsets, indicating that their difficulty is not solely caused by objective-scale misestimation but also by more challenging modeling, reasoning, or search requirements. Moreover, the positive $\Delta$ Acc. values across all datasets suggest that a correct objective-scale estimate consistently benefits performance, while the magnitude of this gap reflects how strongly each dataset depends on the pre-generation prior. These observations support that the objective-scale prior serves as a useful guidance signal rather than an overly restrictive filter.

\begin{table}[H]
\centering
\small
\caption{Pre-generation objective-scale diagnostics. ``GT covered by base scale'' reports the percentage of instances whose ground-truth objective lies within the pre-generated objective range $[L_x,U_x]$. The two conditional accuracy columns report model accuracy separately for instances where the ground-truth objective is inside or outside this range. $\Delta$ Acc. denotes the accuracy gap between these two subsets.}
\label{tab:pregeneration-analysis}
\setlength{\tabcolsep}{7pt}
\renewcommand{\arraystretch}{1.15}
\begin{tabular}{l c c c c c}
\toprule
\rowcolor{ExpHeader}
\textbf{Dataset} 
& \textbf{\#Inst.} 
& \makecell{\textbf{GT covered by}\\\textbf{base scale (\%)}} 
& \makecell{\textbf{Acc. when GT}\\\textbf{in scale (\%)}} 
& \makecell{\textbf{Acc. when GT}\\\textbf{out of scale (\%)}} 
& \makecell{\textbf{$\Delta$ Acc.}\\\textbf{(in--out, pp)}} \\
\midrule
NL4OPT & 245 & \textit{95.5} & \textit{94.0} & \textit{63.6} & \textit{30.4} \\
\rowcolor{ExpBand}
MAMO-Easy & 642 & \textit{86.4} & \textit{95.3} & \textit{88.5} & \textit{6.8} \\
MAMO-Complex & 203 & \textit{82.3} & \textit{54.5} & \textit{52.8} & \textit{1.7} \\
\rowcolor{ExpBand}
IndustryOR & 100 & \textit{79.0} & \textit{53.2} & \textit{42.9} & \textit{10.3} \\
OptMATH & 166 & \textit{70.5} & \textit{35.0} & \textit{26.5} & \textit{8.5} \\
\bottomrule
\end{tabular}
\end{table}

\subsection{Compute-Budget-Controlled Evaluation}
\label{app:compute-budget}
Using the same backbone does not, by itself, equalize test-time computation. We therefore compare inference-time methods on OptMATH under a similar target completion-token allowance, while retaining each method's native search or selection procedure. All runs use Qwen3-4B-Instruct-2507 served by vLLM, the same Gurobi environment, and the same execution timeout. BoN-Consensus samples $N$ candidates in one batch and selects the largest executable objective-value cluster; BoN-ORM scores the same batch with the LLM as an outcome reward model and breaks ties by self-consistency. AutoFormulator retains its hierarchical MCTS, while StarOR uses its standard search and adaptation settings. Table~\ref{tab:compute-budget-controlled} reports accuracy, observed completion-token use, wall time, and method-specific search settings.

\begin{table}[H]
\centering
\footnotesize
\caption{Compute-budget comparison on OptMATH. Accuracy is mean $\pm$ sample standard deviation; completion tokens and wall time are observed averages. Token limits in the setting column are per candidate or rollout.}
\label{tab:compute-budget-controlled}
\setlength{\tabcolsep}{5pt}
\renewcommand{\arraystretch}{1.12}
\begin{tabularx}{\columnwidth}{l c c c >{\raggedright\arraybackslash}X}
\toprule
\rowcolor{ExpHeader}
\textbf{Method} & \textbf{Acc. (\%)} & \makecell{\textbf{Tokens}\\\textbf{(K)}} & \makecell{\textbf{Wall time}\\\textbf{(s)}} & \textbf{Setting} \\
\midrule
BoN-Consensus & $30.72\pm1.59$ & 265.0 & 119.76 & $N=60$, 8K max \\
\rowcolor{ExpBand}
BoN-ORM & $28.51\pm1.94$ & 279.9 & 125.51 & $N=60$, 8K max \\
AutoFormulator & $13.86\pm1.20$ & 129.4 & 317.15 & $N=10$, $T=12$, 8K max \\
\rowcolor{ExpAccent}
\textbf{StarOR} & $\mathbf{33.53\pm1.25}$ & 117.4 & 209.1 & $K=8$, $T=10$, 6K max \\
\bottomrule
\end{tabularx}
\end{table}

StarOR attains the highest accuracy in this comparison, improving on BoN-Consensus by $2.81$ percentage points while using fewer observed completion tokens (117.4K versus 265.0K). BoN-ORM reaches $28.51\%$, below BoN-Consensus's $30.72\%$, and AutoFormulator reaches $13.86\%$ with a longer average wall time of 317.15 seconds. StarOR's 209.1-second average wall time nevertheless exceeds BoN-Consensus's 119.76 seconds: sequential search and GRPO updates add latency beyond generation and execution. These results support an accuracy advantage under the tested budget settings, while the differing realized token counts and update costs preclude a claim of identical total compute or lower cost in every deployment.

\subsection{More Ablation Studies}
\paragraph{Four-stage optimization decomposition.}
The four stages follow the dependency structure of an optimization model: problem type and sets establish the modeling type and index domains; parameters and variables are defined over those domains; the objective and constraints use the resulting symbols; and code turns the completed formulation into an executable solver program. This grouping keeps tightly coupled decisions together while exposing the main error-propagation boundaries to tree search and rollout-based feedback. To examine its effect, we compare one-stage (formulation and code), four-stage (StarOR), and six-stage (type and sets, parameters, variables, objective, constraints, code) variants on OptMATH using the same backbone, with stage-appropriate generation instructions. The settings use comparable nominal completion-token caps of 480K, 480K, and 432K, respectively; Table~\ref{tab:stage-decomposition-ablation} also reports their observed token use. Four stages achieve $33.53\pm1.25\%$ accuracy, exceeding one stage by $1.60$ percentage points and six stages by $2.41$ percentage points. The results are consistent with coarse decisions obscuring local credit assignment in one stage and additional dependencies amplifying error propagation in six stages. Because actual token use differs across variants, this comparison supports the chosen granularity under the tested settings without establishing it as uniquely optimal.

\begin{table}[H]
\centering
\footnotesize
\caption{Stage-decomposition ablation on OptMATH. Accuracy is mean $\pm$ standard deviation; tokens are average completed tokens. Search settings are maximum iterations $\times$ parallel rollouts per iteration $\times$ maximum tokens per rollout.}
\label{tab:stage-decomposition-ablation}
\setlength{\tabcolsep}{7pt}
\renewcommand{\arraystretch}{1.12}
\begin{tabular}{l c c c}
\toprule
\rowcolor{ExpHeader}
\textbf{Variant} & \textbf{Acc. (\%)} & \textbf{Tokens (K)} & \textbf{Search setting} \\
\midrule
One-stage & $31.93\pm3.35$ & 123.2 & $4\times10\times12$K \\
\rowcolor{ExpAccent}
\textbf{Four-stage (StarOR)} & $\mathbf{33.53\pm1.25}$ & 117.4 & $10\times8\times6$K \\
Six-stage & $31.12\pm0.92$ & 183.6 & $18\times6\times4$K \\
\bottomrule
\end{tabular}
\end{table}

\paragraph{Final Repair.}
We also evaluate the effect of the final repair budget after terminal candidate selection. For execution errors, the error-repair prompt uses the failure trace to correct Gurobi code, targeting issues such as syntax errors, missing imports, undefined variables, and solver-status output. For infeasible outcomes, a separate prompt requests an infeasibility diagnosis, an improved mathematical model, and corresponding corrected code. The latter may therefore change the formulation implemented by the final code, although it does not rerun the four-stage search. Table~\ref{tab:repair-ablation-optmath} reports the OptMATH comparison across repair budgets. The default setting allows two repair rounds; additional rounds can improve accuracy but incur more wall time.

\begin{table}[H]
\centering
\footnotesize
\caption{Final repair ablation on OptMATH. ``Repair rounds'' denotes the maximum number of final repair attempts after terminal candidate selection.}
\label{tab:repair-ablation-optmath}
\setlength{\tabcolsep}{7pt}
\renewcommand{\arraystretch}{1.15}
\begin{tabular}{l c c c}
\toprule
\rowcolor{ExpHeader}
\textbf{Variant} & \makecell{\textbf{Repair}\\\textbf{rounds}} & \textbf{Acc. (\%)} & \makecell{\textbf{Time}\\\textbf{Cost (s)}} \\
\midrule
\rowcolor{ExpAccent}
\textbf{StarOR (Ours)} & \textbf{2} & \textbf{32.5} & \textbf{205.2} \\
\midrule
repair = 0 & 0 & 30.7 & 194.6s \\
\rowcolor{ExpBand}
repair = 4 & 4 & 33.1 & 219.8s \\
repair = 8 & 8 & 33.1 &  252.1s\\
\bottomrule
\end{tabular}
\end{table}

\section{Computational Cost and Efficiency Analysis}
\label{app:cost}

\subsection{Execution Budget and Computational Overhead}
\label{app:cost-main}

StarOR is designed for optimization modeling settings where a small formulation error can invalidate a downstream decision. Its runtime is therefore best understood as a structured test-time optimization cost rather than as a raw count of generated programs. In this section, we decompose the wall-clock time into the components that are actually sequential in our implementation: reward-prior construction, batched rollout generation, node-level GRPO update, and parallel execution-based reward evaluation.

\paragraph{Wall-clock decomposition.}
For a problem instance that runs for $I$ MCTS iterations with expansion size $K$, the per-instance wall-clock time can be approximated as
\begin{equation}
    T_{\textsc{StarOR}} \approx T_{\mathrm{reward\text{-}prior}} +
    \sum_{t=1}^{I}\left[
    T_{\mathrm{rollout}}(K) + T_{\mathrm{GRPO}}(K) + T_{\mathrm{exec}}^{\parallel}(K,N_t)
    \right] + T_{\mathrm{repair}},
\end{equation}
where $T_{\mathrm{rollout}}(K)$ denotes the batched generation and rollout-to-code completion time for the $K$ sibling candidates, $T_{\mathrm{GRPO}}(K)$ denotes the transient LoRA update time, and $T_{\mathrm{exec}}^{\parallel}(K,N_t)$ denotes the wall-clock execution time for the original instance and the $N_t$ test cases. The execution term is parallelized across candidates and cases: although each candidate is protected by a $30$s timeout, a normal execution takes about $1$s on average, so execution contributes a small batch-level overhead instead of a serial $K(1+N_t)$ multiplier. Consequently, the main sequential cost is the number of search/adaptation rounds.

\begin{table}[H]
\centering
\scriptsize
\caption{Component-level wall-clock estimate for StarOR with $K=8$ on OptMATH. The estimate matches the measured time scale in Table~\ref{tab:scaling}: the observed average time is $205.2$s per sample.}
\label{tab:appendix-time-decomposition}
\setlength{\tabcolsep}{3pt}
\renewcommand{\arraystretch}{1.15}
\begin{tabularx}{\columnwidth}{@{} l c X @{}}
\toprule
\textbf{Component} & \textbf{Approx. time} & \textbf{Explanation} \\
\midrule
Reward-prior construction & $\approx 4$s & Roughly two single-rollout equivalents for objective-scale estimation and test-case preparation \\
Rollout generation per iteration & $\approx 22$s & Batched generation and completion for $K=8$ sibling candidates. \\
GRPO update per iteration & $\approx 7$s & Node-level LoRA backpropagation and policy update. \\
Parallel execution per iteration & $\approx 1$s & Original and test case executions are run in parallel; the $30$s solver limit is a timeout cap. \\
Average number of iterations & $\approx 7.1$ & Empirical mean number of MCTS/adaptation rounds on OptMATH. \\
\bottomrule
\end{tabularx}
\end{table}

\paragraph{Consistency with empirical runtime.}
Plugging these values into the decomposition gives
\begin{equation}
    T_{\textsc{StarOR}}(K=8)
    \approx 4 + 7.1 \times (22 + 7 + 1)
    \approx 217\text{s}.
\end{equation}
This simple estimate is intentionally conservative because it treats all iterations as full iterations with both rollout and GRPO update. In practice, some terminal or low-utility iterations are shorter, several callback operations overlap with batched rollout execution, and optional repair is not triggered for every sample. These effects reduce the observed average to $205.2$s in Table~\ref{tab:scaling}, which is close to the theoretical estimate. The decomposition therefore supports the runtime interpretation of StarOR: the dominant cost is not serial solver execution, but repeated rollout-and-adaptation rounds that refine the instance-specific policy during search.

\paragraph{Compute efficiency.}
The key question is not which method uses the smallest absolute test-time budget, but which method converts test-time compute into correct formulations more effectively. The scaling results in Table~\ref{tab:scaling} support this view: StarOR with $K=4$ reaches $25.9\%$ accuracy on OptMATH in $93.9$s, compared with Best-of-$N$ at $N=32$ ($24.9\%$) in $104.2$s. This indicates that the additional computation in StarOR is not merely spent on independent resampling; it is reused to update the instance-specific policy and steer later search toward more reliable formulation decisions.

\subsection{Difficulty-Aware Resource Allocation}
\label{app:cost-difficulty}

A core strength of StarOR is its ability to adaptively allocate computational resources based on problem complexity. This difficulty-awareness is reflected in the search depth, convergence speed of the reward signals, and the final wall-clock time:

\paragraph{Case 1: Low-Complexity Instances (e.g., NL4OPT).} 
Problems in the NL4OPT benchmark typically feature direct linear constraints and clear objective mappings. For these instances, StarOR often achieves high semantic consensus ($r_{\text{sem}}$) and structural agreement ($r_{\text{struct}}$) within the first $2$--$3$ iterations. Consequently, the MCTS logic triggers early termination, keeping the average time-per-sample minimal.

\paragraph{Case 2: High-Complexity Instances (e.g., OptMATH).} 
In contrast, OptMATH contains dense numerical descriptions and intricate logical dependencies. Early search nodes often yield diverging objective values (low $r_{\text{sem}}$) or inconsistent variable types (low $r_{\text{struct}}$). This lack of consensus prevents premature termination and drives the model to utilize the full MCTS budget and more GRPO update steps to resolve modeling ambiguities.

Table~\ref{tab:appendix-runtime-stats} illustrates this adaptive behavior by comparing representative statistics across easy and complex benchmarks. The table should be read together with the cost decomposition above: the average time is driven primarily by the number of sequential search/adaptation rounds, while parallel code execution contributes only a small batch-level overhead.

\begin{table}[H]
\centering
\scriptsize
\caption{Difficulty-aware runtime statistics. StarOR ($K=8$) spends little time on benchmarks where semantic and structural consensus emerges early, and allocates more iterations to datasets with ambiguous variables, dense constraints, or unstable objective consensus.}
\label{tab:appendix-runtime-stats}
\setlength{\tabcolsep}{3.5pt}
\renewcommand{\arraystretch}{1.15}
\begin{tabularx}{\columnwidth}{@{} l c c c X @{}}
\toprule
\textbf{Dataset} & \textbf{Difficulty} & \textbf{Avg. Iter.} & \textbf{Avg. Time} & \textbf{Dominant runtime behavior} \\
\midrule
\rowcolor{gray!5} NL4OPT & Easy & $\sim 3.1$ & 65.1s & Early semantic consensus; most instances terminate after shallow stage-wise exploration. \\
MAMO-Easy & Easy & $\sim 3.2$ & 70.4s & Simple LP/MILP structure; extra time mainly comes from code completion and verification. \\
\midrule
MAMO-Complex & Complex & $\sim 6.5$ & 188.5s & Ambiguous variable/constraint mapping; more rounds are needed before structural agreement. \\
\rowcolor{gray!5} OptMATH & Complex & $\sim 7.1$ & 205.2s & Dense numeric descriptions and unstable objective consensus; search often uses a deeper budget. \\
\bottomrule
\end{tabularx}
\end{table}

These statistics support the intended deployment pattern of StarOR. On easier instances, the reward signals converge quickly and the method behaves like a lightweight verifier around a small number of search rounds. On harder instances, the additional cost is spent precisely where one-shot and Best-of-$N$ are weakest: resolving variable definitions, constraint directions, and objective-scale disagreements through repeated search-and-adaptation. Thus, the runtime increase is difficulty-aware rather than uniformly applied to every problem.
\newpage

\section{Implementation Details}
\label{app:details}
Experiments are conducted on a single NVIDIA H20 140GB GPU using the veRL \citep{sheng2024hybridflow} framework for RL training.

\paragraph{GRPO.}
StarOR performs a lightweight online GRPO update after each non-terminal MCTS expansion based verl Implementation.  For a selected node at stage $s$, the current policy consists of the frozen backbone parameters $\phi$ and a transient LoRA adapter $\Delta\phi$, yielding the stage-conditioned policy $\pi_{\phi+\Delta\phi}(\cdot \mid x,\tau_{\le s-1})$.  The adapter is initialized at the beginning of each problem instance and is reset after the instance finishes, so no test-time update is carried across benchmark examples.

At an expansion step, we sample a sibling group of $K$ continuations from the same prefix,
\begin{equation}
    z_i^{(s)} \sim \pi_{\phi+\Delta\phi_{\mathrm{old}}}
    \bigl(\cdot \mid x,\tau_{\le s-1}\bigr),
    \qquad i=1,\ldots,K,
\end{equation}
and roll each continuation to executable code for reward evaluation.  Let $R_i$ denote the scalar multi-faceted reward assigned to the $i$-th sibling after combining semantic consensus, execution, test-case robustness, and structural consistency.  We use the siblings as a local comparison group and compute the GRPO advantage without a learned value function:
\begin{equation}
    \mu_R = \frac{1}{K}\sum_{j=1}^{K}R_j,\qquad
    \sigma_R = \operatorname{std}_{j=1}^{K}(R_j),
    \qquad
    A_i = \frac{R_i-\mu_R}{\sigma_R+\epsilon},
    \label{eq:appendix-grpo-adv}
\end{equation}
where $\epsilon$ is a small numerical constant.  In implementation, the scalar reward is placed on the final valid response token and then summed over the response, so Eq.~\eqref{eq:appendix-grpo-adv} is exactly the group-normalized outcome advantage.  The scalar $A_i$ is broadcast to the valid response tokens selected by the response mask:
\begin{equation}
    A_{i,t}=A_i\,m_{i,t},
\end{equation}
where $m_{i,t}\in\{0,1\}$ selects valid response tokens from the beginning of the generated response up to, but not including, the \texttt{<python>} block. Under the prescribed response format, this prefix contains the \texttt{<thought>} reasoning and the current-stage formulation; the completed code provides evaluation feedback but is excluded from the update.

The LoRA adapter is then updated with a PPO-style clipped policy loss.  Let $y_{i,t}$ be the $t$-th token of the generated response and define the token-level importance ratio
\begin{equation}
    \rho_{i,t}(\Delta\phi)=
    \exp\left[
    \log \pi_{\phi+\Delta\phi}(y_{i,t}\mid x,\tau_{\le s-1},y_{i,<t})
    -
    \log \pi_{\phi+\Delta\phi_{\mathrm{old}}}(y_{i,t}\mid x,\tau_{\le s-1},y_{i,<t})
    \right].
\end{equation}
With clipping range $\varepsilon_{\mathrm{clip}}$, the clipped surrogate for each valid token is
\begin{equation}
    \ell^{\mathrm{clip}}_{i,t}(\Delta\phi)
    =
    -\min\left(
    \rho_{i,t}(\Delta\phi) A_{i,t},
    \operatorname{clip}\bigl(\rho_{i,t}(\Delta\phi),1-\varepsilon_{\mathrm{clip}},1+\varepsilon_{\mathrm{clip}}\bigr) A_{i,t}
    \right).
\end{equation}
Our implementation follows the veRL dual-clip variant for negative advantages, using an additional constant $c>1$ to cap overly large negative-advantage losses:
\begin{equation}
    \ell^{\mathrm{pg}}_{i,t}(\Delta\phi)
    =
    \begin{cases}
    \ell^{\mathrm{clip}}_{i,t}(\Delta\phi), & A_{i,t}\ge 0,\\[2mm]
    \min\!\left(\ell^{\mathrm{clip}}_{i,t}(\Delta\phi), -cA_{i,t}\right), & A_{i,t}<0.
    \end{cases}
\end{equation}

In addition to PPO clipping, we regularize the update toward the reference policy.  We do not subtract the KL term from the reward in our default configuration; instead, we add an actor-level KL loss with coefficient $\beta$.  Using the low-variance KL estimator, the per-token regularizer is
\begin{equation}
    \begin{aligned}
    d^{\mathrm{KL}}_{i,t}(\Delta\phi)
    &=
    \exp\!\left(\delta_{i,t}\right)-\delta_{i,t}-1,\\
    \delta_{i,t}
    &=
    \log\pi_{\mathrm{ref}}(y_{i,t}\mid\cdot)
    -\log\pi_{\phi+\Delta\phi}(y_{i,t}\mid\cdot).
    \end{aligned}
\end{equation}
The final optimization objective for one sibling group is therefore
\begin{equation}
    \mathcal{L}_{\mathrm{GRPO}}(\Delta\phi)
    =
    \frac{1}{\sum_{i,t}m_{i,t}}
    \sum_{i=1}^{K}\sum_t
    m_{i,t}
    \left[
    \ell^{\mathrm{pg}}_{i,t}(\Delta\phi)
    +
    \beta\, d^{\mathrm{KL}}_{i,t}(\Delta\phi)
    \right],
    \label{eq:appendix-grpo-loss}
\end{equation}
and only the LoRA parameters $\Delta\phi$ are optimized.  After the update, the refreshed adapter is synchronized to the rollout engine, so subsequent MCTS expansions sample from the instance-adapted policy $\pi_{\phi+\Delta\phi}$.

\paragraph{Test-case.}
The robustness test cases are prepared once before the MCTS search for each problem instance.  StarOR first converts the raw problem into a numeric instance representation and a feature catalog.  This catalog is constructed by deterministic parsing rather than by free-form generation: inline numerical mentions in the natural-language description are stored as keys such as \texttt{num\_0}, while numerical entries extracted from markdown-style tables are stored as table keys.  Each entry records its numeric value, source type, local text snippet, and a heuristic importance score.  Scores are higher when the surrounding text or table header contains OR-relevant terms such as \emph{cost}, \emph{profit}, \emph{capacity}, \emph{demand}, \emph{budget}, \emph{limit}, or relational language.  The ranked entries are exposed to the planner as a compact feature catalog
\begin{equation}
    \mathcal{F}_x=\{(F_m, k_m, v_m, q_m, h_m)\}_{m=1}^{M},
\end{equation}
where $F_m$ is a stable feature id (e.g., \texttt{F01}), $k_m$ is the internal instance key, $v_m$ is the original numeric value, $q_m$ is the source, and $h_m$ is the supporting snippet.

Given $\mathcal{F}_x$, StarOR uses an auxiliary pre-search generation step to produce two structured artifacts: (i) a conservative objective-scale envelope for the original instance, and (ii) $N_t$ robustness tests.  The LLM is asked to return tagged JSON blocks rather than executable code.  A test case contains a case id, a list of coordinated patches, an objective-scale envelope, and a short rationale:
\begin{equation}
    c_j=\bigl(\mathrm{id}_j,\{(F_m,\mathrm{op},a)\}_{m\in \mathcal{P}_j},
    \mathrm{scale}_j,\mathrm{rationale}_j\bigr).
\end{equation}
The patch list is interpreted programmatically.  Each patch must reference an existing feature id in $\mathcal{F}_x$; the implementation maps the id back to its internal numeric key $k_m$ and applies the edit to a deep copy of the original instance.  Three edit forms are supported:
\begin{equation}
    v'_m =
    \begin{cases}
    a, & \mathrm{op}=\texttt{replace},\\
    a\,v_m, & \mathrm{op}=\texttt{scale},\\
    v_m+a, & \mathrm{op}=\texttt{shift}.
    \end{cases}
\end{equation}
Integer-valued features are rounded back to integers after editing.  Invalid patches, such as unknown feature ids or edits to non-numeric fields, are discarded.  The resulting perturbed instance stores the normalized patch metadata and is cached in the problem instance under the precomputed test-case field.

Importantly, the LLM is not called again when computing $r_{\mathrm{test}}$ during MCTS.  During search, every candidate program is executed directly on the precomputed perturbed instances, in parallel when multiple tests are available.  The observed objective is then checked against the corresponding case-specific objective-scale envelope.  Thus, the test-case reward is an execution-time verification signal: LLM reasoning is used only once to design the stress tests before search, while the per-candidate reward computation is deterministic given the generated code, the cached perturbation patches, and the objective-scale filters.  If structured test-case precomputation fails, the current veRL pipeline marks the test-case reward as disabled for that instance rather than repeatedly querying the LLM inside the search loop.

\subsection{Hyperparameters}
\label{app:hyper}
\label{appendix:hyperparameters}

This section provides a comprehensive summary of the hyperparameters and implementation configurations for StarOR. Table~\ref{tab:appendix-hyperparameters} details the settings for the backbone model, Monte Carlo Tree Search (MCTS), and the Group Relative Policy Optimization (GRPO) components.
\begin{table}[H]
\centering
\footnotesize
\caption{Main hyperparameters for StarOR.}
\label{tab:appendix-hyperparameters}
\renewcommand{\arraystretch}{1.1}
\begin{tabularx}{\textwidth}{@{} ll X @{}}
\toprule
\textbf{Category} & \textbf{Parameter} & \textbf{Value / Setting} \\
\midrule
\textbf{Backbone} & Model & Qwen3-4B-Instruct-2507 \\
                  & Max Response Length & 6,144 tokens \\
                  & Sampling & Temperature 1.0, Top-$p=0.95$ \\
\midrule
\textbf{Search (MCTS)} & Budget & Max iterations $T=10$ \\
                       & Expansion Group Size & $K=8$ siblings per node \\
                       & PUCT Constant & $c_{\mathrm{puct}}=1.414$ \\
                       & Prior Softmax & Temperature $\eta=0.7$ \\
                       & Backpropagation & Decay factor $\rho=0.95$ for clusters \\
                       & Stop & one-shot suppression factor $\gamma_{\mathrm{sup}}=0.5$ \\
                       & Clustering & tolerance $\epsilon_c=0.01\%$ \\
\midrule
\textbf{Adaptation (RL)} & Optimizer & GRPO (Online) \\
                         & Learning Rate & $1 \times 10^{-4}$ \\
                         & LoRA Config & Rank 8, Alpha 16, all linear layers \\
                         & KL Penalty & Coefficient $0.001$, low-variance estimator \\
\midrule
\textbf{Rewards} & Reward Components & $r_{sem}$: Semantic, $r_{exec}$: Exec, $r_{test}$: Test-case, $r_{struct}$: Structural \\
                 & Phase 1 (Iter 1--3) & $(0.2, 0.5, 0.2, 0.1)$ for $r_{sem}$, $r_{exec}$, $r_{test}$, $r_{struct}$   \\
                 & Phase 2 (Iter 4--5) & $(0.4, 0.4, 0.1, 0.1)$ \\
                 & Phase 3 (Iter > 5) & $(0.6, 0.2, 0.1, 0.1)$ \\
                 & Out-of-scale Penalty & Multiplier $\lambda=0.5$ for $r_{sem}$ and $r_{test}$ \\
                 & Test-Case Number & $N_t=3$ \\
\midrule
\textbf{Others} & Consensus Tolerance & Relative objective tolerance $0.01\%$ \\
                    & Code Refinement & Max 2 repair rounds at termination \\
                    & Solver Backend & Gurobi, 30s limit per execution \\
\bottomrule
\end{tabularx}
\end{table}

\subsection{Details of Reward System}
\label{app:reward}
The reward system is designed for the unlabeled test-time setting, where ground-truth answers are unavailable during search. For candidate $i$, it combines four weak but complementary signals:
\begin{equation}
R_i = \max\left(0, w_{\mathrm{sem}} r_{\mathrm{sem},i}
    + w_{\mathrm{exec}} r_{\mathrm{exec},i}
    + w_{\mathrm{test}} r_{\mathrm{test},i}
    + w_{\mathrm{struct}} r_{\mathrm{struct},i}\right),
\end{equation}

\paragraph{Execution reward $r_{exec}$.}
The execution reward is a binary indicator of whether the generated Python/Gurobi program executes without fatal errors. Specifically, the implementation treats any program free of syntax and runtime exceptions as an effective success, prioritizing the model's executability. This broad definition includes not only cases that yield recognizable Gurobi optimality messages or parseable objective values but also any run that completes its process normally. Conversely, only timeouts, unhandled code exceptions, and completely missing solver outputs are treated as failures.

\paragraph{Semantic reward $r_{\mathrm{sem}}$.}
Following the definition in Eq.~\ref{reward_sem}, StarOR calculates the semantic consistency reward for each candidate $i$ based on the size of its objective-value cluster $|C_{\mathrm{sem}, i}|$. Valid objective values are clustered using a relative tolerance of $0.01\%$. To ensure numerical stability during test-time optimization, the practical implementation employs a smoothed version of the consensus ratio:
\begin{equation}
    r_{\mathrm{sem}, i} = \frac{|C_{\mathrm{sem}, i}| + \alpha_1}{K_{\mathrm{valid}} + \alpha_1 \cdot \max(N_{\mathrm{clusters}}, K_{\min})} \cdot \lambda,
\end{equation}
where $\alpha_1=0.6$ is the smoothing coefficient, $K_{\min}=3$ is the baseline cluster scale, $K_{\mathrm{valid}}$ is the number of executable candidates, and $N_{\mathrm{clusters}}$ denotes the total number of identified clusters in the rollout group. The multiplier $\lambda$ corresponds to the \textbf{objective-scale penalty} described in Section~\ref{reward}: $\lambda = 1.0$ if the objective value $O_i$ falls within the predicted scale $[L_x, U_x]$, and $\lambda = 0.5$ otherwise. This reward design ensures that independently generated formulations reaching the same plausible objective value receive higher credit.

\paragraph{Structural consistency reward $r_{struct}$.}
\label{app:reward-struct}
Following the definition in Eq.~\ref{reward_struct}, the structural reward $r_{\text{struct}, i}$ evaluates candidate $i$ by its architectural consistency within the group. To ensure the reward is bounded in $[0, 1]$, the $\text{Norm}(\cdot)$ operator is implemented as a dimension-wise average. Specifically, let $\mathbf{v}_i$ be the signature vector containing the counts of 5 core components:
\begin{equation}
    \mathbf{v}_i = [n_{\mathrm{bin}}, n_{\mathrm{int}}, n_{\mathrm{cont}}, n_{\mathrm{con}}, \mathrm{sense}].
\end{equation}
The implementation incorporates a smoothing term $(\alpha_4, K_4)$ into the consensus ratio for each dimension $d$, and the final reward is calculated as:
\begin{equation}
    r_{\text{struct}, i} = \frac{1}{5} \sum_{d \in \mathbf{v}_i} \sqrt{\frac{|C(i, d)| + \alpha_4}{K_{\mathrm{valid}} + \alpha_4 \cdot K_4}},
\end{equation}
where $|C(i, d)|$ is the number of candidates sharing the same value on dimension $d$, $K_{\mathrm{valid}}$ is the count of executable candidates, $\alpha_4 = 0.4$ is the smoothing coefficient, and $K_4 = 3$ is the normalization constant. This formulation ensures that when perfect consensus is reached ($|C| = K_{\mathrm{valid}}$), the reward approaches 1.0 (subject to smoothing), effectively rewarding structurally stable modeling trajectories.

\paragraph{Test-case reward $r_{test}$.}
Following the definition in Eq.~\ref{reward_test}, the test-case reward $r_{\text{test}, i}$ evaluates the robustness of candidate $i$ against synthetic test cases. For each problem instance, StarOR precomputes $N_t=3$ perturbed versions (e.g., modified numeric capacities or coefficients) with associated objective-scale envelopes. Candidate $i$ is executed on each perturbed instance, and its performance on case $j$ is assigned a score $S_{i,j}$ based on the following criteria:
\begin{itemize}
    \item $S_{i,j} = 1.0$: Execution succeeds, and the objective value falls within the case-specific scale.
    \item $S_{i,j} = \lambda$ ($\lambda = 0.5$): Execution succeeds, but the objective value is outside the predicted scale.
    \item $S_{i,j} = 0.0$: Execution fails (e.g., syntax error, timeout, or infeasibility).
\end{itemize}
The final test-case reward is the average across all test cases:
\begin{equation}
    r_{\text{test}, i} = \frac{1}{N_t} \sum_{j=1}^{N_t} S_{i,j}, \quad S_{i,j} \in \{0, \lambda, 1\}.
\end{equation}
By incorporating the soft-penalty factor $\lambda$, this design ensures that out-of-scale but executable programs are penalized rather than entirely discarded, acknowledging the potential conservative nature of the initial objective-scale estimation.

\paragraph{Objective-scale Prior Estimation.}
Prior to the hierarchical search, StarOR leverages the base policy to estimate a conservative objective-scale envelope for the problem instance. The estimator receives only the problem statement and parsed numerical features; reference answers and ground-truth objective values are not supplied to scale generation, reward computation, branch selection, early stopping, or LoRA updates. This prior acts as a grounded reference for the reward system, particularly for the $\lambda$-penalty in $r_{\text{sem}}$ and $r_{\text{test}}$. The estimated scale is structured as a JSON object, enabling programmatic verification across multiple dimensions:
\begin{center}
\begin{minipage}{0.85\textwidth} 
\begin{lstlisting}[
    language=json,
    basicstyle=\small\ttfamily, % 这里改成了 \small，如果觉得太大可以改为 \footnotesize
    backgroundcolor=\color{gray!3!white},
    frame=single,
    frameround=tttt, % 可选：让边框圆角化，看起来更现代
    rulecolor=\color{gray!30},
    keywordstyle=\color{blue},
    stringstyle=\color{teal},
    breaklines=true,
    caption={Example for objective-scale priors.}
]
{
  "kind": "interval",
  "lower": 500,
  "upper": 2800,
  "sign_relation": "positive",
  "magnitude": {
    "min_order": 2, 
    "max_order": 4, 
    "use_abs": true
  },
  "reject_exact": [0]
}
\end{lstlisting}
\end{minipage}
\end{center}

The runtime verification engine applies a hierarchical filtering logic: 
\begin{enumerate}[leftmargin=*, label=\arabic*)]
    \item \textbf{Validity Check:} Rejects non-finite objectives (e.g., NaN, $\pm\infty$) and values (e.g. $0$) specified in the \texttt{reject\_exact} list.
    \item \textbf{Structural Constraints:} Enforces sign consistency (e.g., non-negativity) and magnitude-order constraints (e.g., ensuring the value is within $10^2$ to $10^4$).
    \item \textbf{Interval Grounding:} Validates the objective against explicit numeric bounds. 
\end{enumerate}
To accommodate the potential conservatism of the zero-shot backbone, we apply a $10\%$ margin relaxation to the interval bounds during final selection and early stopping. This buffer aims to reduce false negatives when the LLM returns rounded or approximate endpoints. The $10\%$ margin and out-of-scale multiplier $\lambda=0.5$ were fixed before benchmark evaluation as simple heuristic choices, without fine-grained or benchmark-specific tuning; we do not claim that either value is optimal.

\paragraph{Test-case Generation.}
To synthesize the $N_t$ test cases for robustness evaluation, the \textbf{test-case generator} first constructs a compact numeric feature catalog by parsing the natural-language problem description. Each numerical value or table entry is assigned a unique identifier (e.g., \texttt{F01}, \texttt{F02}), with priority given to features proximal to critical OR semantics such as \emph{cost}, \emph{profit}, \emph{capacity}, \emph{demand}, \emph{budget}, and \emph{limit}. Leveraging this catalog and original problem description, the generator produces coordinated patches targeting high-ranking features simultaneously, accompanied by a re-estimated objective-scale envelope for each generated test case. We show the example in the List.~\ref{list:test}

\begin{center} \label{list:test}
\begin{minipage}{0.85\textwidth} 
\begin{lstlisting}[
    language=json,
    basicstyle=\small\ttfamily, % 这里改成了 \small，如果觉得太大可以改为 \footnotesize
    backgroundcolor=\color{gray!3!white},
    frame=single,
    frameround=tttt, % 可选：让边框圆角化，看起来更现代
    rulecolor=\color{gray!30},
    keywordstyle=\color{blue},
    stringstyle=\color{teal},
    breaklines=true,
    caption={Example for test-case generation.}
]
{
    {
    "case_id":"easier_route_key_edges_down",
    "patches":[
        {"fid":"F04","new_value":38},
        {"fid":"F08","new_value":24},
        {"fid":"F05","new_value":42}
        ],
    "obj_scale":{
        "kind":"interval",
        "lower":180,
        "upper":245,
        "sign_relation":"positive",
        "magnitude":{
            "min_order":2,
            "max_order":3,
            "use_abs":true
            },
        "reject_exact":[0]
        },
    },
}
\end{lstlisting}
\end{minipage}
\end{center}

\paragraph{Prompt templates.}
The two templates below are used before search to construct the objective-scale prior and the robustness tests. They are displayed in a verbatim-style prompt box, so the raw prompt can be pasted directly into the paper source without manually inserting LaTeX line breaks or escaping prompt placeholders.

\begin{starorprompt}{Base Objective-Scale Prompt Template}
You are an OR objective-scale analyst.
Return exactly two tagged blocks and nothing else:
<analysis>...</analysis>
<base_scale>{...JSON...}</base_scale>

Goal:
Produce a conservative runtime filter(obj, scale) for the ORIGINAL optimization instance.

Priority:
Prioritize correctness over precision. The interval should be wide enough to include all mathematically plausible optimal objective values, but narrow enough to reject absurd, dimensionally impossible, or semantically invalid values.

Analysis logic in <analysis>:
1. Identify the optimization sense (minimize or maximize) and the physical meaning of the objective.
2. Estimate a theoretical floor and ceiling from the numeric data, such as total demand, maximum capacity, fixed costs, largest unit costs, or profit bounds.
3. Explain why the selected scale is safe and should not reject the true optimum.
4. Explicitly exclude impossible values, such as negative costs when all costs are nonnegative, zero objective values when fixed costs are mandatory, or magnitudes that violate the instance scale.

Output schema for <base_scale>:
{
  "kind": "interval",
  "lower": number,
  "upper": number,
  "sign_relation": "positive | nonnegative | negative | mixed | unknown",
  "magnitude": {
    "min_order": integer,
    "max_order": integer,
    "use_abs": true
  },
  "reject_exact": [number, ...]
}

Task description:
{{task_description}}

Numeric snapshot:
{{compact_instance_json}}

Feature catalog:
{{feature_catalog_json}}

Think carefully in <analysis>, then provide the final JSON object in <base_scale>.
\end{starorprompt}

\begin{starorprompt}{Perturbation Test-Case Generation Prompt Template}
You are an expert Operations Research (OR) stress-test engineer.
Return exactly two tagged blocks and nothing else:
<analysis>...</analysis>
<tests>[...JSON list...]</tests>

Goal:
Design robustness tests that challenge the formulation's sensitivity, feasibility boundaries, and objective-scale consistency.

Each test should contain:
- case_id: a short descriptive identifier.
- patches: 1-3 coordinated feature edits using feature ids from the catalog.
- obj_scale: a conservative objective-scale envelope for the perturbed instance.
- rationale: a short explanation of why this perturbation is meaningful.

Analysis logic in <analysis>:
1. Identify sensitive parameters whose changes strongly affect feasibility or objective value, such as bottleneck capacities, high-cost coefficients, demands, budgets, resource limits, or service requirements.
2. Plan 3-5 realistic scenarios, such as resource scarcity, relaxed capacity, extreme cost variation, demand surge, or boundary feasibility.
3. For each scenario, estimate a safe objective range that is broad enough to include plausible optimal values but still excludes impossible values.

Requirements for <tests>:
- Use only feature ids that appear in the feature catalog.
- Prefer small, meaningful perturbations over random large changes.
- Preserve unit consistency and problem semantics.
- Avoid patches that make the natural-language problem nonsensical unless the scenario explicitly tests infeasibility handling.
- The obj_scale must use the schema:
  {
    "kind": "interval",
    "lower": number,
    "upper": number,
    "sign_relation": "positive | nonnegative | negative | mixed | unknown",
    "magnitude": {"min_order": integer, "max_order": integer, "use_abs": true},
    "reject_exact": [number, ...]
  }

Task description:
{{task_description}}

Numeric snapshot:
{{compact_instance_json}}

Feature catalog:
{{feature_catalog_json}}

Think carefully in <analysis>, then provide the final JSON list in <tests>.
\end{starorprompt}

\subsection{Algorithm}
Algorithm~\ref{alg:StarOR} summarizes the complete StarOR inference procedure.

\begin{algorithm}[htbp]
\caption{StarOR: Synergistic Tree Search and Test-Time Policy Adaptation}
\label{alg:StarOR}
\begin{algorithmic}[1]
\REQUIRE Problem instance $x$, base policy $\pi_{\phi}$, search budget $T$, expansion size $K$
\ENSURE Final executable solver program $c^\star$
\STATE Construct the pre-generation prior: objective-scale envelope $[L_x,U_x]$ and synthetic test cases $\mathcal{D}_{\mathrm{test}}$
\STATE Initialize root node $n_0$, terminal candidate set $\mathcal{C}\leftarrow\emptyset$, stage archives $\{\mathcal{B}_s\}_{s=1}^{3}$, and transient adapter $\Delta\phi\leftarrow 0$
\STATE Initialize the one-step suppression mask $\mathcal{S}\leftarrow\emptyset$
\FOR{$t=1,\dots,T$}
\STATE Select an expandable leaf $n$ using the PUCT score under the current policy $\pi_{\phi+\Delta\phi}$ and suppression mask $\mathcal{S}$
\STATE Let $\tau_{\le s-1}$ be the partial formulation stored at $n$, where $s$ is the next stage to be generated
\IF{$s=\texttt{code}$ and the one-time code deferral has not been used}
\STATE Add the current trajectory and its objective-consensus cluster to $\mathcal{S}$ for one iteration
\STATE Mark the code deferral as used and \textbf{continue}
\ENDIF
\STATE Sample a sibling group $\{z_i^{(s)}\}_{i=1}^{K}\sim\pi_{\phi+\Delta\phi}(\cdot\mid x,\tau_{\le s-1},s)$ and complete it into an executable rollout $c_i$
\STATE Execute each $c_i$ on $x$ and $\mathcal{D}_{\mathrm{test}}$; compute $r_{\mathrm{sem}}$, $r_{\mathrm{exec}}$, $r_{\mathrm{test}}$, $r_{\mathrm{struct}}$, and total reward $R_i$
\STATE Cluster executable siblings by objective consensus and structural signatures; let $\mathcal{I}$ be the evaluated rollout index set
\STATE Add children $\{n_i\}_{i\in\mathcal{I}}$ to the tree and assign priors from normalized model log-likelihoods
\FOR{each child $n_i$}
\STATE Store its partial trajectory, executable rollout, cluster ID, and individual reward $R_i$
\ENDFOR
\STATE Compute the rollout-group value $\bar{R}=\frac{1}{|\mathcal{I}|}\sum_{i\in\mathcal{I}}R_i$
\STATE Backpropagate $\bar{R}$ along the selected path from $n$ to $n_0$
\STATE Apply discounted group backpropagation to same-cluster sibling nodes using decay $\rho^{\ell}$
\IF{$s \neq \texttt{code}$}
\STATE Estimate GRPO advantages from $\{R_i\}_{i\in\mathcal{I}}$ and update the transient adapter $\Delta\phi$
\STATE Add the scored partial trajectories $\{\tau_{\le s,i}\}_{i\in\mathcal{I}}$ to archive $\mathcal{B}_s$
\ELSE
\STATE Add terminal rollouts $\{c_i\}_{i\in\mathcal{I}}$ to $\mathcal{C}$
\STATE Select $c^\star$ by objective consensus, objective-scale filtering, and optional repair
\STATE \textbf{break}
\ENDIF
\IF{objective consensus is stable across recent distinct stages}
\STATE Complete the consensus-supported trajectory into code, add it to $\mathcal{C}$, and select $c^\star$ with the terminal-selection rule
\STATE \textbf{break}
\ENDIF
\STATE Clear expired entries in the one-step suppression mask $\mathcal{S}$
\ENDFOR
\IF{$c^\star$ has not been selected}
\STATE Select $c^\star$ from $\mathcal{C}$ or archived executable rollouts by global objective-consensus voting and objective-scale-aware tie breaking
\ENDIF
\STATE \textbf{return} $c^\star$
\end{algorithmic}
\end{algorithm}
\newpage

\section{Prompt Templates and Structured Instructions}
\label{app:prompts}

This section reports the structured prompting interface used by StarOR. 

\subsection{Global System Instruction}

Every model query is prefixed with the following system instruction. The same instruction is used for all four formulation stages, auto-completion rollouts, and local policy adaptation samples.

\begin{starorprompt}{System Instruction for StarOR}
You are a helpful assistant with expertise in operations research modeling and the Gurobi Python solver.

Think step by step before producing the requested tagged output.

Output only clean tag-specific content inside the required tags. Do not include extra explanations outside the tags.

At the end of each rollout, you must provide complete executable Gurobi Python code inside <python>...</python>.
\end{starorprompt}




\subsection{Stage Rollout Templates}

\begin{starorprompt}{Stage 1: Type and Sets}
You are a professional optimization problem analyst. Your task is to extract the problem type and the minimum necessary indexing sets from a natural-language optimization problem.

Optimization problem:
{{task_description}}

Instructions:
1. Think step by step inside <thought>. Identify the decision context, objective direction, resources, agents, items, time periods, locations, or other entities that need indexing.
2. Output the problem type inside <Type>. Include the optimization family when identifiable, such as LP, MILP, assignment, transportation, facility location, routing, scheduling, blending, or production planning.
3. Output the indexing sets inside <Sets>. Define only the sets needed for a clean mathematical formulation.
4. After completing this stage, continue the remaining formulation and provide complete executable Gurobi Python code inside <python>.

Required output order:
<thought>...</thought>
<Type>...</Type>
<Sets>...</Sets>
<python>...</python>

MANDATORY FORMAT RULES:
1. <Type> should summarize:
- optimization type: LP / MILP / NLP / MINLP and so on.
- classical OR family when identifiable: TSP / Facility Location Problem / VRP (Vehicle Routing Problem) and so on. 
- Explanation: Provide a brief sentence outlining the rationale and key points.

2. <Sets> should define the minimum necessary indexing sets.
- set_name: description: {elements if explicitly enumerable}
Example:
- s: Employee types: {f,p} where f=full-time workers, p=part-time workers


Here is some code example:
<python>
import gurobipy as gp
from gurobipy import GRB

# Create model
......(here is core modeling code)

model.optimize()

status = model.status
if status == GRB.OPTIMAL:
    optimal = model.objVal
    print(f"Optimal value: {{optimal}}")
else:
    print(f"Model status: {{status}}")
</python>
\end{starorprompt}

\begin{starorprompt}{Stage 2: Parameters and Variables}
You are a professional optimization problem analyst. Your task is to define the numerical parameters and decision variables given the previously committed problem type and sets.

Optimization problem:
{{task_description}}

Committed Type and Sets:
{{type_sets_content}}

Instructions:
1. Think step by step inside <thought>. Identify all numeric constants, tables, capacities, demands, costs, profits, budgets, bounds, and logical coefficients required by the model.
2. Output the parameters inside <Parameters>. Preserve units and map each parameter to the relevant set indices.
3. Output the decision variables inside <Variables>. Specify variable domain, index set, and semantic meaning.
4. After completing this stage, continue the remaining formulation and provide complete executable Gurobi Python code inside <python>.

Required output order:
<thought>...</thought>
<Parameters>...</Parameters>
<Variables>...</Variables>
<python>...</python>

MANDATORY FORMAT RULES:
1. Parameters format:
- Indexed parameter:
  - param_index: description [unit][indexed by set_name] (data type): value_or_semantic_value
- Global parameter:
  - param: description [unit] (data type): value_or_semantic_value

2. Variables format:
- Indexed variable:
  - x_index: description (domain)
- Global variable:
  - x: description (domain)

3. Naming rules:
- parameter names must be concise and consistent with sets/entities.
- variable names must be concise and consistent with later symbolic modeling.
- use the same terminology as previous stages.
- do not rename entities casually.

Here is some code example:
<python>
import gurobipy as gp
from gurobipy import GRB

# Create model
......(here is core modeling code)

model.optimize()

status = model.status
if status == GRB.OPTIMAL:
    optimal = model.objVal
    print(f"Optimal value: {{optimal}}")
else:
    print(f"Model status: {{status}}")
</python>
\end{starorprompt}

\begin{starorprompt}{Stage 3: Objective and Constraints}
You are a professional optimization problem analyst. Your task is to write the mathematical objective and constraints using the previously committed type, sets, parameters, and variables.

Optimization problem:
{{task_description}}

Committed Type and Sets:
{{type_sets_content}}

Committed Parameters and Variables:
{{para_var_content}}

Instructions:
1. Think step by step inside <thought>. Determine the objective sense, the exact objective expression, and every feasibility condition stated or implied by the problem.
2. Output the objective inside <Objective>. Include whether the problem is a minimization or maximization problem.
3. Output the constraints inside <Constraints>. Group constraints by semantic role, such as demand satisfaction, capacity, assignment, budget, balance, precedence, compatibility, or domain restrictions.
4. After completing this stage, provide complete executable Gurobi Python code inside <python>.

Required output order:
<thought>...</thought>
<Objective>...</Objective>
<Constraints>...</Constraints>
<python>...</python>

MANDATORY FORMAT RULES:
1. Objective format:
- objective_name: description: $LaTeX expression$

2. Constraints format:
- constraint_name: description: $LaTeX expression$ (type: Equality/Inequality)

3. All symbols in objective/constraints must come from previous stages.
4. Use symbolic parameters rather than hard-coded numeric coefficients whenever possible.
5. Output results directly. Do NOT output chain-of-thought.

CONSISTENCY RULES:
- Every variable appearing in the objective must be defined earlier.
- Every parameter appearing in the objective must be defined earlier.
- Every symbol in every constraint must be defined earlier.
- Objective and constraints must together reflect the original task faithfully.
- Do not omit key structural constraints.
- Do not add assumptions that materially change the problem.

Here is some code example:
<python>
import gurobipy as gp
from gurobipy import GRB

# Create model
......(here is core modeling code)

model.optimize()

status = model.status
if status == GRB.OPTIMAL:
    optimal = model.objVal
    print(f"Optimal value: {{optimal}}")
else:
    print(f"Model status: {{status}}")
</python>
\end{starorprompt}

\begin{starorprompt}{Stage 4: Code}
You are a professional optimization model implementer. Your task is to faithfully translate the committed formulation into executable Gurobi Python code.

Optimization problem:
{{task_description}}

Committed Type and Sets:
{{type_sets_content}}

Committed Parameters and Variables:
{{para_var_content}}

Committed Objective and Constraints:
{{obj_con_content}}

Instructions:
1. Think briefly inside <thought>. Check that the committed formulation is internally consistent.
2. Output only the executable Gurobi Python implementation inside <python>.
3. Translate the committed formulation faithfully. Do not redesign the model unless the committed formulation contains an obvious contradiction that would prevent execution.

Required output order:
<thought>...</thought>
<python>...</python>

Here is some code example:
<python>
import gurobipy as gp
from gurobipy import GRB

# Create model
......(here is core modeling code)

model.optimize()

status = model.status
if status == GRB.OPTIMAL:
    optimal = model.objVal
    print(f"Optimal value: {{optimal}}")
else:
    print(f"Model status: {{status}}")
</python>

\end{starorprompt}

\subsection{Repair and Completion Templates}

When final candidate is syntactically no-objective, StarOR invokes a lightweight repair prompt. If code is error, we use the error repair prompt; if model is infeasible, we use the infeasible repair prompt.

\begin{starorprompt}{Code Error Repair Prompt}
You are an experienced operations research algorithm engineer. You are presented with an operations research problem and a previous attempt to model and code a solution. That attempt resulted in an error.
Problem Description:
{task_description}

Previous Code Solution Attempt:
<python>
{code_text}
</python>

After running the provided code from the previous attempt, the following error occurred:
{error_info}

Your task:
Based on the information above, please perform the following:
1. Analyze Root Cause & Identify Pitfalls 
- Thoroughly analyze the root cause of the error.
- Summarize potential pitfalls or common mistakes related to this type of code error.
2. Provide Corrected Gurobi Code:
- Write the complete and corrected Python code using the 'gurobipy' library to accurately solve the problem.

Please structure your response strictly as follows:
## Cause of the Error and Potential Pitfalls:
<thought> (Your detailed analysis of the error's cause and a summary of potential pitfalls.) </thought>
## Corrected Gurobi Code:
<python>
import gurobipy as gp
from gurobipy import GRB

# Create model
......(here is core modeling code)

model.optimize()

status = model.status
if status == GRB.OPTIMAL:
    optimal = model.objVal
    print(f"Optimal value: {{optimal}}")
else:
    print(f"Model status: {{status}}")
</python>
Please think step by step.
\end{starorprompt}

\begin{starorprompt}{Code Infeasible Repair Prompt}
You are an experienced operations research algorithm engineer. You are presented with an operations research problem and a previous attempt to model and code a solution. That attempt resulted in an error.


Task description:
{task_description}

Fixed model blocks (unchanged):
{model_text}

Current code:
<python>
{code_text}
</python>

Execution error:
{execution_text}

You are an experienced operations research algorithm engineer. You are presented with an operations research problem and a previous attempt to model and code a solution. That attempt resulted in an infeasible solution.
Problem Description:
{task_description}

Previous Model:
{model_text}

Code Solution Attempt:
<python>
{code_text}
</python>

After running the provided code from the previous attempt, the answer could not provide a feasible solution.

Your task:
Based on the information above, please perform the following:

1. Analyze Root Cause & Identify Pitfalls
- Thoroughly analyze the root cause of the infeasibility.
- Summarize potential pitfalls or common mistakes related to this type of infeasibility.

2. Provide an Improved Mathematical Model: 
- Develop a mathematical model for correctly
- modeling this OR problem. This should address the flaws in the previous attempt.

3. Provide Corrected Gurobi Code:
Write the complete and corrected Python code associated with the mathematical model using the 'gurobipy' library to accurately solve the problem.

Please structure your response strictly as follows:
## Cause of the Error and Potential Pitfalls:
<thought> (Your detailed analysis of the error's cause and a summary of potential pitfalls.) </thought>

## Corrected Mathematical Model:
<Type>
[Identify the problem class: LP, MILP, NLP, etc.]
</Type>

<Sets>
[Define all indices and sets with clear descriptions]
</Sets>

<Parameters>
[Define all constants and data structures, including units]
</Parameters>

<Variables>
[Define decision variables, their domains (Binary, Non-negative, etc.), and physical meanings]
</Variables>

<Objective>
[Mathematical expression of the objective function with Max/Min direction]
</Objective>

<Constraints>
[List all mathematical constraints. Ensure they are indexed correctly (e.g., $ \forall i \in I$) and clearly explained]
</Constraints>

## Corrected Gurobi Code:
<python>
import gurobipy as gp
from gurobipy import GRB

# Create model
......(here is core modeling code)

model.optimize()

status = model.status
if status == GRB.OPTIMAL:
    optimal = model.objVal
    print(f"Optimal value: {{optimal}}")
else:
    print(f"Model status: {{status}}")
</python>
Note: Do not rewrite the model from scratch; instead, surgically patch the existing model and code by addressing valid feedback while critically filtering out any incorrect or redundant signals.
Please think step by step.
\end{starorprompt}

\section{Licenses for Existing Assets}
\label{app:licenses}

We use existing datasets, models, software frameworks, and solver assets only for academic research and benchmarking. Table~\ref{tab:appendix-licenses} summarizes the main assets used in our experiments, their original creators or maintainers, and the license or terms that we identified from the public release pages. Regarding benchmark preparation, we follow the data cleaning and standardization protocols established by \citet{chen2025sirl}. 

\begin{table}[H]
\centering
\scriptsize
\setlength{\tabcolsep}{3pt}
\renewcommand{\arraystretch}{1.15}
\caption{Existing assets used in StarOR. ``Terms'' denotes a non-open-source usage agreement rather than an OSI-style license.}
\label{tab:appendix-licenses}
\begin{tabularx}{\columnwidth}{p{0.18\columnwidth}p{0.24\columnwidth}p{0.18\columnwidth}X}
\toprule
\textbf{Asset} & \textbf{Creator / Maintainer} & \textbf{License / Terms} & \textbf{Use in this paper} \\
\midrule
Qwen3-4B-Instruct-2507 & Qwen team / Alibaba Cloud & Apache-2.0 & Backbone model for all StarOR and same-backbone test-time baselines. Model card: \url{https://huggingface.co/Qwen/Qwen3-4B-Instruct-2507}. \\
\midrule
verl & Volcano Engine / ByteDance Seed and community & Apache-2.0 & RL training and rollout infrastructure adapted for our TTRL-OR runtime. Repository: \url{https://github.com/volcengine/verl}. \\
\midrule
Gurobi Optimizer & Gurobi Optimization, LLC & Proprietary Gurobi license; academic licenses are available for non-commercial academic research & Solver backend for executing generated Python optimization models. We use Gurobi under academic/research terms and cite the Gurobi reference manual. \\
\midrule
NL4OPT & NL4OPT organizers; converted optimal-answer version by Cardinal Operations / ORLM & Official competition repository: MIT; converted Hugging Face dataset card: CC BY-NC 4.0 & Evaluation benchmark for natural-language LP modeling. We cite the NL4OPT competition and use the converted/checked benchmark only for non-commercial academic evaluation. \\
\midrule
MAMO-Easy and MAMO-Complex & MAMO authors; public mirror by Cardinal Operations & Public Hugging Face dataset card: CC BY-NC 4.0 & Evaluation benchmarks for easy and complex LP/MILP modeling. We use cleaned versions that preserve source attribution and benchmark identity. \\
\midrule
IndustryOR & ORLM / Cardinal Operations & Public Hugging Face dataset card: CC BY-NC 4.0; ORLM code repository: Apache-2.0 & Industrial OR evaluation benchmark. We use the 100-instance benchmark for non-commercial research evaluation. \\
\midrule
OptMATH-Bench & OptMATH authors & Publicly released for research; no separate dataset license was identified on the project page at the time of writing & Evaluation benchmark for difficult optimization modeling. We cite the OptMATH paper/project page and use only the released benchmark instances for academic evaluation. \\
\bottomrule
\end{tabularx}
\end{table}

\paragraph{External services and closed models.}
Some baselines report results from proprietary or externally hosted models such as GPT-4, DeepSeek-V3.1, and DeepSeek-R1. For these systems, we cite the corresponding technical reports or model papers and report results either from prior work or from API-based evaluation under the providers' terms of service. These models are not redistributed with our artifacts.

\section{Broader Impacts}
\label{app:broader-impacts}

StarOR aims to make optimization modeling more accessible by reducing the effort required to translate natural-language decision problems into executable solver programs. This can benefit domains such as logistics, manufacturing, scheduling, energy planning, healthcare operations, and public-sector resource allocation, where better optimization models can improve utilization, reduce waste, and support more transparent decision-making. The framework is especially useful when problem descriptions vary across instances and collecting large supervised datasets for every domain is impractical.

At the same time, automated optimization modeling carries operational and societal risks. A generated formulation may be executable but semantically misaligned with the intended decision problem, leading to recommendations that violate constraints, omit stakeholder preferences, or optimize an incomplete objective. These risks are more serious in high-stakes settings such as healthcare scheduling, disaster response, infrastructure planning, and workforce allocation. StarOR reduces this risk through staged formulation, solver execution, structural checks, perturbation tests, and objective-consensus rewards, but these mechanisms are safeguards rather than correctness guarantees.

The method also increases test-time computation. Additional compute can improve reliability, but it has economic and environmental costs. Our results suggest that this compute is most valuable for difficult instances where one-shot decoding is unreliable; practical deployments should therefore use difficulty-aware budgets, early stopping, and smaller expansion groups for easy instances. We recommend using StarOR as a decision-support assistant: generated formulations, solver traces, reward diagnostics, and final-selection evidence should be logged and reviewed by domain experts before decisions are enacted.

\end{document}